\documentclass{article}
\usepackage{ijcai26}

\usepackage{times}
\usepackage{soul}
\usepackage{url}
\usepackage[hidelinks]{hyperref}
\usepackage[utf8]{inputenc}
\usepackage[small]{caption}
\usepackage{graphicx}
\usepackage{amsmath}
\usepackage{amsthm}
\usepackage{booktabs}
\usepackage{algorithm}
\usepackage{algorithmic}
\usepackage[switch]{lineno}
\usepackage{amssymb}
\usepackage{multirow}

\usepackage{graphicx}          
\usepackage{setspace}          
\usepackage[most]{tcolorbox}   
\usepackage{enumitem}          
\usepackage{caption}           
\usepackage{amsmath}           

\usepackage{etoc} 

\usepackage{tabularx} 

\usepackage{colortbl} 
\usepackage{tabularx}
\usepackage[table]{xcolor} 
\usepackage{booktabs}
\usepackage{float}
\usepackage{tcolorbox}

\definecolor{thinkcolor}{RGB}{0, 0, 255}     
\definecolor{searchcolor}{RGB}{0, 128, 128}  
\definecolor{infocolor}{RGB}{204, 102, 0}    
\definecolor{answercolor}{RGB}{192, 0, 0}    

\newcommand{\think}[1]{\textcolor{thinkcolor}{\textless think\textgreater} #1 \textcolor{thinkcolor}{\textless /think\textgreater}}
\newcommand{\search}[1]{\textcolor{searchcolor}{\textless search\textgreater} #1 \textcolor{searchcolor}{\textless /search\textgreater}}
\newcommand{\info}[1]{\textcolor{infocolor}{\textless information\textgreater} #1 \textcolor{infocolor}{\textless /information\textgreater}}
\newcommand{\answer}[1]{\textcolor{answercolor}{\textless answer\textgreater} #1 \textcolor{answercolor}{\textless /answer\textgreater}}

\newcommand{\doc}[1]{\textcolor{infocolor}{\textless doc\textgreater} #1 \textcolor{infocolor}{\textless /doc\textgreater}}

\newcommand{\doctype}[1]{\textcolor{infocolor}{\textless doc-type\textgreater} #1 \textcolor{infocolor}{\textless /doc-type\textgreater}}

\newcommand{\subquestion}[1]{\textcolor{searchcolor}{\textless sub-question\textgreater} #1 \textcolor{searchcolor}{\textless /sub-question\textgreater}}

\newcommand{\subanswer}[1]{\textcolor{answercolor}{\textless sub-answer\textgreater} #1 \textcolor{answercolor}{\textless /sub-answer\textgreater}}

\newcommand{\mainanswer}[1]{\textcolor{answercolor}{\textless main-answer\textgreater} #1 \textcolor{answercolor}{\textless /main-answer\textgreater}}

\newcommand{\ie}{\textit{i.e.,}~}

\usepackage{nth}
\usepackage{relsize}

\title{ARMOR: Adaptive Curriculum Meta-Learning \\ 
for Noise-Robust RAG Reasoning
}

\author{
Yan Wang$^{1,2}$\thanks{Equal contributors.} 
\and
Yuxin Zhang$^{1,2}$\footnotemark[1]\and
Shenyu Zhang$^{1,2}$\and
Yongrui Chen$^{1,2}$\and
Sheng Bi$^3$\and
Guilin Qi$^{1,2}$\thanks{Corresponding author.} \\
\affiliations
$^1$School of Computer Science and Engineering, Southeast University, China\\
$^2$Laboratory of New Generation Artificial Intelligence Technology and Its Interdisciplinary Applications (Southeast University), Ministry of Education, China\\
$^3$School of Law, Southeast University, China\\
\emails
\{yanwangstu, zzyx\_cs, gqi\}@seu.edu.cn\\
}

\begin{document}
\etocdepthtag.toc{mtoc}
\maketitle

\begin{abstract}
    Retrieval-Augmented Generation (RAG) systems have demonstrated remarkable effectiveness in mitigating hallucinations by incorporating external knowledge. 
    However, the retrieval process inevitably introduces noise, posing significant challenges to RAG robustness.
    Fundamentally, noise robustness is a composite competency encompassing noise identification, retrieval filtering, and robust reasoning.
    Unfortunately, existing methods often fail to capture this holistic nature, typically restricted to single-dimensional enhancements or hampered by noise-induced optimization interference.
    To address these limitations, we decompose RAG robustness into three sub-tasks: knowledge boundary perception, retrieval verification, and robust reasoning.
    Accordingly, we propose ARMOR, an \underline{A}daptive Cu\underline{r}riculum \underline{M}eta-Learning for N\underline{o}ise-\underline{R}obust RAG Reasoning. 
    Inspired by the teaching process, ARMOR introduces an adaptive task scheduler as a \textit{Teacher} to curate training plans according to the absolute learning progress of RAG generators (\textit{Student}).
    Distinct from GRPO, ARMOR utilizes inter-group advantage estimation during learning to effectively mitigate mode collapse caused by limited output spaces.
    Experiments on three complex reasoning datasets (\textit{2WikiMultiHopQA}, \textit{MusiQue}, and \textit{HotPotQA}) indicate that ARMOR achieves synergistic improvements in retrieval noise robustness, significantly outperforming baselines in output quality.
    Our code and datasets are available at the \href{https://drive.google.com/file/d/1fR5eA3shgKN2q-Savl4OWsj-0r21DsnD/view?usp=sharing}{link}.
\end{abstract}

\section{Introduction}
Retrieval-Augmented Generation (RAG) has emerged as a critical paradigm for mitigating hallucinations in Large Language Models (LLMs) by incorporating external knowledge~\cite{rag1,rag2}.
However, the retrieval process is inherently imperfect and inevitably introduces irrelevant or misleading \textit{Retrieval Noise}, which poses a severe challenge to the robustness of RAG systems~\cite{robust3,robust4}.
Fundamentally, noise robustness is not a standalone trait but a composite capability.
It requires generative systems to simultaneously perform across three distinct dimensions: identifying knowledge boundaries, filtering noise from retrieved context, and executing correct reasoning~\cite{robust1}.

Unfortunately, current methods often focus on specific facets of noise robustness~\cite{robust2}.
Some studies employ supervised fine-tuning to train generators to detect noise and execute robust reasoning within noisy contexts~\cite{trust-rag,instruct-rag}.
Conversely, others leverage generative systems' intrinsic knowledge boundary perception to reduce unnecessary retrieval~\cite{DRAGIN,skr}.
However, these methods generally restrict optimization to isolated capabilities, lacking the synergy required for comprehensive robustness.
Reinforcement Learning (RL) offers a paradigm to address this by enabling the co-evolution of these capabilities via end-to-end policy optimization~\cite{grpo,gigpo}.

Nevertheless, applying standard RL to enhance retrieval noise robustness poses two critical challenges.
First, the reward signals derived from the multi-dimensional optimization objectives of noise robustness may result in mutual interference or cancellation, thereby preventing the generative system from converging to an optimal policy for any specific capability~\cite{R3-RAG}.
Second, these capabilities exhibit varying degrees of complexity and follow an underlying sequential logic.
Just as \textit{a student must learn addition and subtraction before multiplication and division}, the optimization of noise robustness necessitates a sequential learning trajectory.
However, standard RL algorithms typically disregard this inherent progression, attempting to optimize all noise robustness capabilities simultaneously~\cite{RAG-RL}.
This simultaneous learning strategy often induces an overwhelming optimization burden, restricting the generative system from effectively learning these capabilities.

To address these challenges, we propose ARMOR, an \underline{A}daptive Cu\underline{r}riculum \underline{M}eta-Learning for N\underline{o}ise-\underline{R}obust RAG Reasoning.
ARMOR leverages curriculum-based Reinforcement Learning to enhance the comprehensive noise robustness of generative systems.
Accordingly, we decouple the composite objective of noise robustness into three sub-tasks: (1) knowledge boundary perception, (2) retrieval verification, and (3) robust reasoning.
Specifically, knowledge boundary perception determines whether the model's intrinsic knowledge suffices or external retrieval is required.
Retrieval verification distinguishes relevant evidence from irrelevant noise within the retrieved context.
Robust reasoning integrates valid information to generate correct answers even in noisy contexts.
Systematically acquiring these capabilities is essential for establishing a resilient defense against retrieval noise.

We draw inspiration from the human teaching process, where teachers dynamically adjust the curriculum based on the student's learning feedback~\cite{ALP}.
ARMOR models the training process as an adaptive curriculum meta-learning framework, and introduces an adaptive task scheduler as a \textit{teacher}.
In each training step, the teacher dynamically selects a specific sub-task for the student (the RAG generator), leveraging Absolute Learning Progress (ALP) to quantify the student's real-time proficiency and prioritize the capability with the highest potential for improvement.

By prioritizing the most critical capability at each stage, ARMOR allows the optimizer to focus on a single dimension of robustness, effectively avoiding the mutual interference and cancellation of reward signals.
To further stabilize this decoupled training, we introduce inter-group advantage estimation.
Distinct from standard group relative policy optimization (GRPO)~\cite{grpo}, inter-group advantage estimation extends advantage estimation to a batch-level global scope.
By combining inter-group advantage estimation with the adaptive curriculum meta-learning framework, ARMOR effectively eliminates task heterogeneity between different inputs and mitigates mode collapse stemming from limited output spaces.
Experiments on three complex reasoning datasets (\textit{2WikiMultiHopQA}~\cite{2wiki}, \textit{MusiQue}~\cite{musique}, and \textit{HotPotQA}~\cite{hotpotqa}) indicate 
that ARMOR achieves synergistic improvements in retrieval noise robustness, significantly outperforming baselines in output quality.

We summarize our contributions as follows:
\begin{itemize}
    \item We propose ARMOR, utilizing an ALP-based adaptive task scheduler to dynamically adjust the learning curriculum based on the student's feedback.
    \item We introduce inter-group advantage estimation to stabilize training and mitigate mode collapse stemming from limited output spaces.
    \item Experiments on three noise-injected complex reasoning datasets demonstrate that ARMOR achieves synergistic improvements in retrieval noise robustness.
\end{itemize}

\section{Related Work}
\paragraph{Noise Robustness in RAG.}
Enhancing the robustness of RAG systems against retrieval noise has become a pivotal research direction. 
A primary defense against noise involves explicit mechanisms to evaluate and filter retrieved documents before or during generation.
\textsc{CRAG}~\cite{crag} employs a lightweight retrieval evaluator to assess document quality, triggering corrective actions when the confidence is low.
Similarly, \textsc{TrustRAG}~\cite{trust-rag} utilizes clustering and self-assessment to identify and filter out adversarial or misleading information.
Rather than filtering inputs, \textsc{Chain-of-Note}~\cite{chain-of-note} trains models to generate sequential reading notes, explicitly assessing document pertinence before formulating an answer. 
\textsc{InstructRAG}~\cite{instruct-rag} further advances this by instructing models to synthesize rationales that explain how the answer is derived from noisy contexts.
Additionally, \textsc{RAAT}~\cite{raat} employs adaptive adversarial training, exposing the model to various noise types to internalize robust reasoning capabilities.
\textsc{SKR}~\cite{skr} utilizes the model's self-knowledge to classify questions as ``known'' or ``unknown'' before invoking a retriever.
Similarly, \textsc{DRAGIN}~\cite{DRAGIN} introduces Real-time Information Needs Detection to trigger retrieval only when the model exhibits uncertainty during generation.
However, existing methods typically address noise robustness as an isolated capability by focusing solely on filtering, reasoning, or retrieval triggering.

\paragraph{RL for Robust RAG.}
Recent advancements have increasingly leveraged RL to enhance the robustness of RAG systems against retrieval noise.
Some studies focus on optimizing the interaction policy between reasoning and retrieval.
For instance, \textsc{Search-R1}~\cite{search-r1} and \textsc{R1-Searcher++}~\cite{search++} utilize outcome-based RL to train models to autonomously search queries within reasoning steps, enabling dynamic knowledge acquisition.
To enhance noise filtering, \textsc{R3-RAG}~\cite{R3-RAG} employs a relevance-based process reward to verify retrieved documents at each step.
Similarly, \textsc{LeTS}~\cite{lets} integrates knowledge redundancy and matching rewards to penalize irrelevant retrieval and prevent error propagation in multi-hop reasoning.
\textsc{EviNote-RAG}~\cite{evinote} restructures the workflow into a ``retrieve-note-answer'' paradigm, using an evidence quality reward to filter noise and distill key evidence before generation.
Furthermore, \textsc{RAG-RL}~\cite{RAG-RL} investigates static curriculum strategies to improve model robustness against distractors.
In contrast, ARMOR decomposes robustness into three distinct sub-tasks, utilizing an ALP-based adaptive task scheduler to dynamically adjust the learning curriculum based on the generator's feedback.

\begin{figure*}
    \centering
    \includegraphics[width=\textwidth]{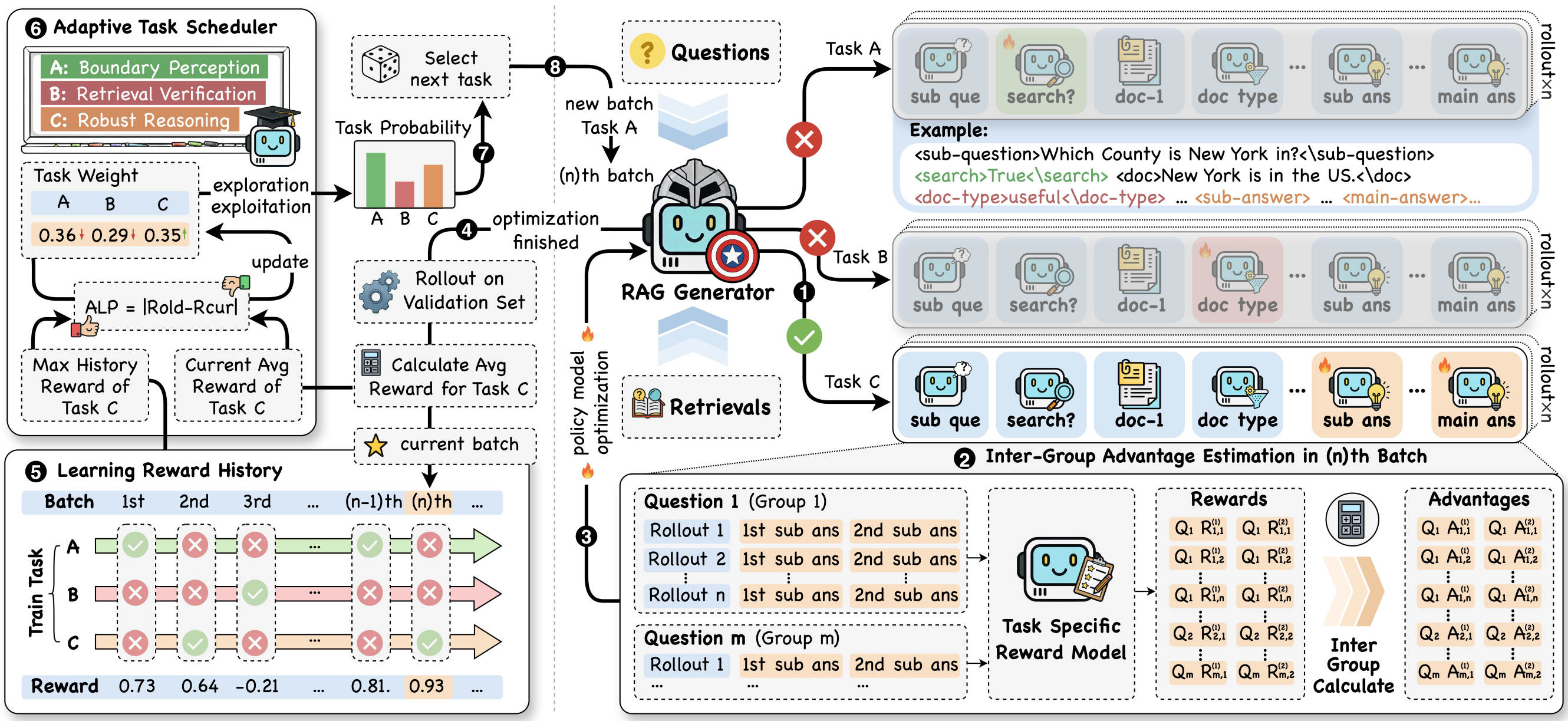}
    \caption{The overall framework of ARMOR. (\textbf{Left}) The Adaptive Task Scheduler functions as a \textit{Teacher}, dynamically prioritizing one of the three robustness sub-tasks (Knowledge Boundary Perception, Retrieval Verification, and Robust Reasoning) for each training batch based on the Absolute Learning Progress (ALP). (\textbf{Right}) The RAG Generator (\textit{Student}) generates multiple rollouts to learn the selected task. (\textbf{Bottom Right}) Inter-Group Advantage Estimation computes reward advantages across the global batch scope rather than within groups, effectively mitigating mode collapse and stabilizing the decoupled training process.}
    \label{fig:Method-Framework}
    \vspace{-1em}
\end{figure*}

\section{Preliminaries}
\paragraph{Task Definition:}
We focus on the task of RAG under noisy retrieval conditions. 
Given a QA pair $(q, a) \in \mathcal{T}_{train}$, the RAG system utilizes a retriever to fetch a set of external documents $\mathcal{D}=\{d_1,d_2,\cdots,d_k\}$ based on the question $q$ from a large corpus.
The retrieved set $\mathcal{D}$ inevitably contains noise due to the inherent imperfections of the retriever. 
We denote the subset of golden documents as $\mathcal{D}^+ \subseteq \mathcal{D}$ and the noisy documents as $\mathcal{D}^- = \mathcal{D} \setminus \mathcal{D}^+$.
The robust RAG generator $\pi_\theta$ aims
to generate the correct answer $a$ by effectively filtering $\mathcal{D}^-$ and reasoning over $\mathcal{D}^+$ in conjunction with its intrinsic knowledge.
We formulate this as maximizing the probability of the correct answer $a$ given the query $q$ and retrieved set $\mathcal{D}$:
\begin{equation}
    \theta^* = \operatorname*{arg\,max}_\theta \sum_{(q, a) \in \mathcal{T}_{train}} \log P_\theta(a \mid q, \mathcal{D}).
\end{equation}

\paragraph{Warm-Up:}
To equip the RAG generator with the fundamental capabilities of problem decomposition and step-by-step reasoning, we employ SFT on reconstructed multi-hop RAG datasets as a warm-up stage. This phase lays the foundation for the subsequent RL phase.
For each pair of \verb|<main-question>| and \verb|<main-answer>|, we construct a reasoning chain to serve as the training target. 
A single reasoning chain consists of structured nodes: \verb|<think>|, \verb|<sub-question>|, \verb|<search>|, \verb|<doc>|, \verb|<doc-type>|, and \verb|<sub-answer>|. 
Specifically, we train the generator to decompose the \verb|<main-question>| into sub-questions, determine retrieval policies via \verb|<search>True<search>| or \verb|<search>False<search>|, verify retrieved documents using \verb|<doc-type>| tags, and generate corresponding sub-answers  based on the reasoning process. 
This structured supervision ensures that the generator explicitly internalizes the ``retrieve-verify-reason" workflow before exploring complex noisy environments. 
Detailed dataset construction procedures are provided in Appendix~\ref{app_sec:Details of Dataset Reconstruction}.

\section{Methodology}

To achieve comprehensive noise robustness, we propose ARMOR, an Adaptive Curriculum Meta-Learning framework. 
As illustrated in Figure~\ref{fig:Method-Framework}, ARMOR consists of two core components: (1) Adaptive Task Scheduler that dynamically curates the learning curriculum based on the learning feedback, and (2) Inter-Group Advantage Estimation mechanism that stabilizes optimization across heterogeneous sub-tasks.

\subsection{Adaptive Curriculum Meta-Learning}
We decouple the composite objective of noise robustness into three sub-tasks: Knowledge Boundary Perception ($\tau_{\text{ret}}$), Retrieval Verification ($\tau_{\text{fil}}$), and Robust Reasoning ($\tau_{\text{ans}}$). 
To avoid the mutual interference of reward signals from distinct tasks, we employ an adaptive task scheduler. 
In each training round $n$, the scheduler selects a task $\tau_s $ from the task set $ \mathcal{T} = \{\tau_{\text{ret}}, \tau_{\text{fil}}, \tau_{\text{ans}}\}$ for optimization, prioritizing tasks with high potential for improvement based on learning feedback.
We initialize the task weights $w_1(\tau) = 1/|\mathcal{T}|$ for all $\tau \in \mathcal{T}$. In the $n$-th round, the selection probability $p_n(\tau)$ is computed to balance exploration and exploitation:

\begin{equation}
    \gamma = \min\left \{  1, \sqrt{\frac{|\mathcal{T}| \ln(|\mathcal{T}|T) }{T} } \right \},
\end{equation}
\begin{equation}
    p_n(\tau) = (1 - \gamma) \frac{w_n(\tau)}{\sum_{\tau' \in \mathcal{T}} w_{n}(\tau')} + \frac{\gamma}{|\mathcal{T}|},
\end{equation}
where $T$ is the predefined training rounds, and $\gamma$ serves as an exploration coefficient to prevent premature convergence to local optima.
The scheduler samples a task $\tau_s \sim p_n(\cdot)$ at the round $n$.
Upon completing the training for the selected task $\tau_s$, we quantify the model's increase in competence using Absolute Learning Progress (ALP)~\cite{ALP}. A higher ALP indicates that the task offers significant potential for performance improvement. The raw ALP is defined as:

\begin{equation}
    ALP(\tau) = 
    \mathbb{I}[\tau = \tau_s] \cdot \bigl\lvert r_n(\tau_s) - r_{\text{old}}(\tau_s) \bigr\rvert,
\end{equation}
where  $r_{n}(\tau_s)$ denotes the task reward received in the current training round $n$~(detailed in section \ref{sec:reward_function}), $r_{\text{old}}(\tau_s)$ represents the maximum historical reward achieved for task $\tau_s$. 
To correct for the bias introduced by probabilistic sampling, we compute the unbiased reward estimation $\hat{r}_n(\tau)$:

\begin{equation}
    \hat{r}_n(\tau) = 
    \frac{ALP(\tau)}{p_n(\tau)},
\end{equation}
detailed proofs are provided in Appendix~\ref{app_sec:Proof of Unbiased Reward Estimation}.
Finally, the scheduler updates the task weights for the next round, incorporating a forgetting factor $\alpha = 1/T$ to smooth updates and retain historical experience:
\begin{equation}
    w_{n+1}(\tau) = w_n(\tau) \cdot \exp\left( \frac{\gamma \hat{r}_n(\tau)}{|\mathcal{T}|} \right) + \frac{e\alpha}{|\mathcal{T}|} \cdot \sum_{\tau' \in \mathcal{T}} w_n(\tau').
\end{equation}

\subsection{Inter-Group Advantage Estimation}
Standard Group Relative Policy Optimization (GRPO) normalizes advantages within a group of rollouts corresponding to a single input. 
However, intra-group normalization results in unstable learning signals for individual tasks during rollouts. 
In our decoupled training setting, the constrained output space induces high similarity among rollouts generated from the same input, which reduces intra-group diversity and exacerbates the risk of mode collapse. 
To overcome this limitation, ARMOR introduces Inter-Group Advantage Estimation, which performs advantage normalization across the entire training batch rather than within isolated inputs.

Given a batch of questions $\mathcal{Q} = \{q_1, \dots, q_m\}$, we generate $l$ rollouts for each question, denoted as:
\begin{equation}
    \mathcal{O}_{\text{train}} = \left\{ o_{i,j} \;\middle|\; 1 \leq i \leq m,\ 1 \leq j \leq l \right\},
\end{equation}
where $ o_{i,j}$ denotes the $j$-th rollout of question $q_i$.
For the selected task $\tau_s$, each rollout $o_{i,j}$ receives a sequence of task-level rewards, denoted as:
\begin{equation}
    \mathbf{r}_{i,j} = \left[ r_{i,j}^{(1)}, \dots, r_{i,j}^{(K_{i,j})} \right]^\top,
\end{equation}
where $r_{i,j}^{\text(k)}$ represents the reward of $k$-th task span in the rollout $o_{i,j}$. In addition, each rollout sequence $o_{i,j}$ is associated with a format reward
 $r_{i,j}^\text{fmt}$, which evaluates whether the rollout sequence conforms to the required format.
We define the global sets of task and format rewards in current batch as:
\begin{align}
    R_{\text{task}} &= \left\{ \mathbf{r}_{i,j} \;\middle|\; 1 \leq i \leq m,\ 1 \leq j \leq l \right\}, \\
    R_{\text{fmt}}  &= \left\{ r_{i,j}^{\text{fmt}} \;\middle|\; 1 \leq i \leq m,\ 1 \leq j \leq l \right\}.
\end{align}

As shown in Figure~\ref{fig:Method-Framework}, we normalize rewards using the global statistics of the entire batch to compute the task-level advantage $A_{i,j}^{\text{task}(k)}$ and format-level advantage $A_{i,j}^{\text{fmt}}$:
\begin{equation}
    A_{i,j}^{\text{task}(k)} = \frac{r_{i,j}^{(k)}-\operatorname{mean}(R_{\text{task}})}{\operatorname{std}(R_{\text{task}})},
\end{equation}
\begin{equation}
    \quad A_{i,j}^\text{fmt} = \frac{r_{i,j}^\text{fmt}-\operatorname{mean}(R_{\text{fmt}})}{\operatorname{std}(R_{\text{fmt}})},
\end{equation}
where $\operatorname{mean}(\cdot)$ and $\operatorname{std}(\cdot)$ compute the mean and standard deviation over the entire batch.
By combining inter-group advantage estimation with the adaptive curriculum meta-learning framework, ARMOR effectively eliminates task heterogeneity between different inputs and provides a more stable training signal.

The final token-level inter-group advantage $A_{i,j}^{(t)}$ combines dual-level signals:
\begin{equation}
    A_{i,j}^{(t)} = \sum_{k=1}^{K_{i,j}} \mathbb{I}[t \in \text{tokens}(i, j, k)] \cdot A_{i,j}^{\text{task}(k)}+\omega A_{i,j}^{\text{fmt}} 
\end{equation}
where $t$ is the token index in rollout $o_{i,j}$, 
$\mathrm{tokens}(i, j, k)$ denotes the token index of the $k$-th task span in rollout $o_{i,j}$.
$A_{i,j}^{(t)}$ represents the token advantage assigned to token $t$ in rollout $o_{i,j}$.
The weighting coefficient $\omega$ controls the relative strength of the format signal.
Then the clipped policy optimization objective of ARMOR is as follows:
\vspace{-1.5em}
\begin{tiny}
\begin{equation}
\begin{aligned}
\mathcal{J}_{\text{ARMOR}}(\theta) & =\mathbb{E}\left[q_i \in \mathcal{Q},\{o_{i,j}\}^l_{j=1} \sim \pi_{\theta_{o l d}}(\cdot  \mid qi)\right]
\frac{1}{\sum_{j=1}^{l}\left|o_{i,j}\right|} \sum_{j=1}^{l}\sum_{t=1}^{\left|o_{i,j}\right|} \\
&\min \bigg[\frac{\pi_{\theta}\left(o_{i,j}^{(t)} \mid q, o_{i,j}^{(<t)}\right)}{\pi_{\theta_{o l d}}\left(o_{i,j}^{(t)} \mid q, o_{i,j}^{(<t)}\right)} A_{i,j}^{(t)}, \\
&\operatorname{clip}\left(\frac{\pi_{\theta}\left(o_{i,j}^{(t)} \mid q, o_{i,j}^{(<t)}\right)}{\pi_{\theta_{o l d}}\left(o_{i,j}^{(t)} \mid q, o_{i,j}^{(<t)}\right)}, 1-\varepsilon_{low}, 1+\varepsilon_{high}\right)  A_{i,j}^{(t)}\bigg].
\end{aligned}
\end{equation}
\end{tiny}

After the training for the task $\tau_s$ in the $n$-th batch, the task reward $r_n(\tau_s) = \text{mean}(R_{\text{task}})$ is calculated on the validation dataset, and returned to the adaptive task scheduler to update the task weights.

\subsection{Reward Modeling} \label{sec:reward_function}
To precisely guide the generator toward robust behaviors, we design tailored reward functions for each sub-task, addressing their specific optimization challenges.

\paragraph{Knowledge Boundary Perception.}
The primary objective of this task is to enable the generator to accurately perceive the boundaries of its intrinsic knowledge.
For each generated sub-question \verb|<sub-question>|, the RAG generator predicts a \verb|<search>| tag to determine whether to invoke the retriever. 
The action space is defined as:
\vspace{-0.4em}
\begin{equation}
    \mathcal{C}_{retriever} = \{\text{true}, \text{false}\}
\end{equation}

To align the model's retrieval behavior with the actual necessity of external information, we design a reward function that evaluates whether the model should retrieve from external knowledge based on the correctness of its direct answer.
We denote the correctness of the answer generated without retrieval as $v_{\text{ans}} \in \{\text{correct}, \text{error}\}$, and the model's chosen retrieve action as $a_{\text{ret}}$. The reward $r_{\text{ret}}$ is defined as follows:
\vspace{-0.5em}
\begin{equation}
    r_{\text{ret}} = 
\begin{cases} 
1 & \text{if } v_{\text{ans}} \text{ is error} \land a_{\text{ret}} = \text{true}, \\
-1 & \text{if } v_{\text{ans}} \text{ is error} \land a_{\text{ret}} = \text{false}, \\
-0.5 & \text{if } v_{\text{ans}} \text{ is correct} \land a_{\text{ret}} = \text{true}, \\
0.5 & \text{if } v_{\text{ans}} \text{ is correct} \land a_{\text{ret}} = \text{false}.
\end{cases}
\end{equation}

\paragraph{Retrieval Verification.}
This task requires the generator to classify each retrieved document into three categories: $\mathcal{C}_{doc} = \{\text{golden}, \text{inconsequential}, \text{irrelevant}\}$ \textcolor{red}~\cite{Magic-Mushroom}.
Standard coarse-grained binary rewards (correct/incorrect classification) often lead to training instability. 
To mitigate this, we propose a fine-grained Retrieval Verification Reward based on the probability margin.
For a document $d$, let $c^*$ be the ground-truth label and $p(c)$ be the predicted probability of class $c$. 
We first calculate the confidence margin $m(d)$ between the target class and the most competitive incorrect class:
\begin{equation}
    m(d) = p(c^*) - \max_{c \neq c^*, c \in \mathcal{C}_{doc}} p(c).
\end{equation}

The final retrieval verification task reward is:
\begin{equation}
    r_\text{fil}(d) = \tanh\big(\beta \cdot m(d)\big),
\end{equation}
where scaling factor $\beta$ is used to control the sensitivity of the reward scaling.

\paragraph{Robust Reasoning.}

The goal of robust reasoning is to generate correct sub-answers \verb|<sub-answer>| and a final answer \verb|<final-answer>| given noisy context.
We compute the answer reward $r_{\text{ans}}$ by first obtaining a raw correctness score $s_{\text{ans}}$ from an LLM evaluator, then linearly mapping it to the range $[-1, 1]$:
\begin{equation}
    r_{\text{ans}} = 2 \cdot \frac{s_{\text{ans}} - s_{\min}}{s_{\max} - s_{\min}} - 1,
\end{equation}
where $[s_{\min}, s_{\max}]$ is the range of correctness score.
Details of the LLM-based evaluation are provided in Appendix~\ref{app_sec:Details of Answer Evaluation}.

\section{Experiment}
In this section, we present empirical evaluations of the effectiveness of ARMOR in defending against retrieval noise across three complex reasoning datasets.
Specifically, we aim to validate the following:
(1) ARMOR's capability in enhancing the noise robustness of RAG generators;
(2) the contribution of core components via ablation studies;
(3) the dynamic trends of learning feedback $p_n(\tau)$ during training.

\subsection{Experiment Setup}
\paragraph{Datasets.}
As detailed in Appendix~\ref{app_sec:Details of Dataset Reconstruction}, we reconstruct \textit{2WikiMultiHopQA} and \textit{MuSiQue} datasets. For each dataset, training set consists of two stages: a warm-up stage and a RL stage. The validation set derives from the reconstructed datasets, while the test set consists of two reconstructed subsets and \textit{HotpotQA}, which serves as an out-of-distribution dataset for evaluating model generalization. The detailed dataset statistics are presented in Table~\ref{tab:dataset-stats}.

\begin{table}[H]
    \centering
    \begin{tabularx}{\linewidth}{l>{\centering\arraybackslash}X>{\centering\arraybackslash}X>{\centering\arraybackslash}X}
        \toprule
        \textbf{Subset} &
        \cellcolor{gray!20}\textit{2Wiki} &
        \cellcolor{gray!20}\textit{MuSiQue} &
        \cellcolor{gray!20}\textit{HotpotQA} \\
        \midrule
        \textbf{Train$_{\text{Warm-Up}}$}    & 4,467   & 496 & -- \\
        \textbf{Train$_{\text{RL}}$}         & 7,716   & 857 & -- \\
        \textbf{Test}       & 1,352   & 375 & 500 \\
        \textbf{Validation} & 1,352   & 375 & -- \\
        \bottomrule
    \end{tabularx}
    \caption{Experiment dataset statistics.}
    \label{tab:dataset-stats}
\end{table}
\vspace{-1em}
\paragraph{Evaluation Metrics.}
We evaluated the model using the following metrics:
(1)~\textbf{Exact Match~(EM)} measures the strict accuracy by determining if the predicted answer matches the ground truth verbatim.
(2)~\textbf{F1-Score~(F1)} captures the token-level overlap between prediction and reference, offering flexibility for answers with minor phrasing variations.
However, these traditional metrics rely on surface-level matching and often fail to capture semantic validity in generative tasks.
To address this, we introduce (3)~\textbf{Correctness~(Cor.)}, an LLM-based metric utilizing \texttt{GPT-5} to assess the factual accuracy and completeness of answers against references.
For clarity, correctness scores are normalized to the range $[0,1]$. 
Detailed evaluation prompts and results aligned with manual annotations are presented in the Appendix \ref{app_sec:Details of Answer Evaluation}.

\paragraph{Baselines.}
We compare ARMOR against several established baselines:
(1)~\textbf{Adaptive Retrieval}: We compare against \textsc{SKR}~\cite{skr} and \textsc{DRAGIN}~\cite{DRAGIN}, which selectively trigger retrieval based on model's intrinsic knowledge boundary to mitigate noise introduction.
(2)~\textbf{Robust Reasoning}: This category includes \textsc{Chain-of-Note}~\cite{chain-of-note} and \textsc{IRCoT}~\cite{ircot}. We employ standard \textsc{CoT} without retrieval as a lower bound to benchmark pure reasoning robustness.
(3)~\textbf{RL-Enhanced RAG}: This includes methods like \textsc{RAG-RL}~\cite{RAG-RL} and \textsc{Search-R1}~\cite{search-r1} that optimize retrieval and reasoning policies via reinforcement learning.
In addition, we include \textsc{No-RAG} (direct inference) and \textsc{Vanilla-RAG} (standard RAG workflow), which serve as performance lower bounds, as they lack specific mechanisms to defend against retrieval noise.
We employ two mainstream LLM series with varying parameter scales as the RAG generators: Llama-3.2-1B, Llama-3.2-3B~\cite{llama3}, Qwen-3-0.6B, Qwen-3-1.7B and Qwen-3-4B~\cite{qwen3}.
For all baselines, we strictly follow the hyperparameter settings and configurations reported in their original studies.
Detailed baseline configurations are provided in Appendix~\ref{app_sec:additional_exp_setup}.

\paragraph{Training Details.} We train the RAG generator for one epoch using Adam optimizer~\cite{Adam} with a learning rate of 5e-5. One epoch includes 857 training rounds, and each round consists of 10 questions.
The clipping coefficient $\varepsilon_{low}$ and $\varepsilon_{high}$ are set to 0.2 and 0.28, separately.
The default rollout number $l$ is set to 2.
The weighting coefficient $\omega$ is set to 1, and the scaling factor $\beta$ in the retrieval verification task is set to 6. 
During the rollout process, we employ \texttt{e5-base-v2}~\cite{e5} as the retriever, retrieving a maximum of three documents per retrieval triggering. For generation, we adopt a sampling temperature of 0.85, along with $\text{top-}p=0.95$ and $\text{top-}k=50$.
All reported results are averaged over three independent runs to ensure statistical reliability.

\begin{table*}[hbtp] \centering
\resizebox{\textwidth}{!}{
\scalebox{0.8}{
\begin{tabular}{lcccccccccccc}
\toprule
\multirow{2}[1]{*}{\textbf{Method}} & \multicolumn{3}{c}{\textit{2WikiMultiHopQA}} & \multicolumn{3}{c}{\textit{MusiQue}} & \multicolumn{3}{c}{\textit{HotpotQA}} & \multicolumn{3}{c}{\textit{Avg.}} \\
\cmidrule(lr){2-4} \cmidrule(lr){5-7} \cmidrule(lr){8-10} \cmidrule(lr){11-13}
& \textbf{Cor.} & \textbf{F1} & \textbf{EM} & \textbf{Cor.} & \textbf{F1} & \textbf{EM} & \textbf{Cor.} & \textbf{F1} & \textbf{EM} & \textbf{Cor.} & \textbf{F1} & \textbf{EM} \\
\midrule
\rowcolor{gray!20}\multicolumn{13}{c}{\textbf{Qwen-3-1.7B}} \\
\textsc{No-RAG} & 37.72 & 37.21 & 32.99 & 3.73 & 2.96 & 0.00 & 24.20 & 22.44 & 15.67 & 21.88 & 20.87 & 16.22 \\
\textsc{Vanilla-RAG} & 47.63 & 44.67 & 38.46 & 24.00 & 27.43 & 17.33 & 45.07 & 45.72 & 38.33 & 38.90 & 39.27 & 31.37 \\
\textsc{CoT} & 39.64 & 17.13 & 11.39 & 8.27 & 6.27 & 1.33 & 22.47 & 18.03 & 13.60 & 23.46 & 13.81 & 8.77 \\
\textsc{CoN} & 63.61 & 13.13 & 4.73 & 29.33 & 12.36 & 5.33 & 54.53 & 22.65 & 17.07 & 49.16 & 16.05 & 9.04 \\
\textsc{IRCoT} & 50.59 & 48.80 & 41.86 & 32.00 & 29.97 & 18.67 & 55.40 & 51.97 & 46.26 & 46.00 & 43.58 & 35.60 \\
\textsc{DRAGIN} & 39.70 & 39.46 & 36.98 & 6.67 & 8.87 & 5.33 & 26.87 & 26.52 & 22.73 & 24.41 & 24.95 & 21.68 \\
\textsc{SKR} & 35.50 & 32.00 & 26.92 & 5.33 & 3.86 & 0.00 & 21.13 & 20.78 & 13.20 & 20.65 & 18.88 & 13.37 \\
\textsc{RAG-RL} & 49.70 & 46.67 & 41.39 & 29.66 & 28.79 & 20.33 & 48.33 & 47.64 & 42.67 & 42.56 & 41.03 & 34.80 \\
\textsc{Search-R1} & 41.57 & 39.61 & 37.28 & 10.67 & 10.55 & 3.73 & 28.27 & 28.21 & 21.93 & 26.84 & 26.12 & 20.98 \\
\textsc{ARMOR}~(warm-up stage) & \cellcolor{cyan!10}70.41 & \cellcolor{cyan!10}70.69 & \cellcolor{cyan!10}68.34 & \cellcolor{cyan!10}38.67 & \cellcolor{cyan!10}39.04 & \cellcolor{cyan!10}29.33 & \cellcolor{cyan!10}47.53 & \cellcolor{cyan!10}45.20 & \cellcolor{cyan!10}38.86 & \cellcolor{cyan!10}52.20 & \cellcolor{cyan!10}51.64 & \cellcolor{cyan!10}45.51 \\
\textsc{ARMOR} & \cellcolor{cyan!25}\textbf{79.14} & \cellcolor{cyan!25}\textbf{79.22} & \cellcolor{cyan!25}\textbf{77.22} & \cellcolor{cyan!25}\textbf{54.67} & \cellcolor{cyan!25}\textbf{57.22} & \cellcolor{cyan!25}\textbf{46.67} & \cellcolor{cyan!25}\textbf{57.73} & \cellcolor{cyan!25}\textbf{54.86} & \cellcolor{cyan!25}\textbf{58.00} & \cellcolor{cyan!25}\textbf{63.85} & \cellcolor{cyan!25}\textbf{63.77} & \cellcolor{cyan!25}\textbf{60.63} \\
\midrule
\rowcolor{gray!20}\multicolumn{13}{c}{\textbf{Llama-3.2-3B}} \\
\textsc{No-RAG} & 28.25 & 30.09 & 25.15 & 5.33 & 2.32 & 0.00 & 22.60 & 17.98 & 13.40 & 18.73 & 16.80 & 12.85 \\
\textsc{Vanilla-RAG} & 59.32 & 57.58 & 53.40 & 16.00 & 17.48 & 10.67 & 50.27 & 44.52 & 36.27 & 41.86 & 39.86 & 33.45 \\
\textsc{CoT} & 36.83 & 17.22 & 14.20 & 8.00 & 3.82 & 1.33 & 35.60 & 23.72 & 18.07 & 26.81 & 14.92 & 11.20 \\
\textsc{CoN} & 51.33 & 7.80 & 1.92 & 14.67 & 3.71 & 0.00 & 47.47 & 10.63 & 6.20 & 37.82 & 7.38 & 2.71 \\
\textsc{IRCoT} & 61.39 & 58.76 & 53.40 & 34.67 & 32.43 & 22.67 & 60.13 & 57.77 & 48.00 & 52.06 & 49.65 & 41.36 \\
\textsc{DRAGIN} & 42.60 & 41.76 & 39.50 & 9.33 & 7.77 & 4.00 & 37.47 & 35.85 & 29.73 & 29.80 & 28.46 & 24.41 \\
\textsc{SKR} & 35.21 & 36.98 & 32.99 & 8.26 & 6.95 & 2.67 & 35.00 & 32.50 & 31.33 & 26.16 & 25.48 & 22.33 \\
\textsc{RAG-RL} & 64.48 & 62.58 & 58.40 & 29.67 & 30.21 & 21.33 & 56.60 & 50.47 & 41.27 & 50.25 & 47.75 & 40.33 \\
\textsc{Search-R1} & 42.31 & 40.76 & 38.76 & 6.67 & 8.50 & 4.00 & 26.33 & 27.22 & 24.53 & 25.10 & 25.49 & 22.43 \\
\textsc{ARMOR}~(warm-up stage) & \cellcolor{cyan!10}71.01 & \cellcolor{cyan!10}71.28 & \cellcolor{cyan!10}68.93 & \cellcolor{cyan!10}32.00 & \cellcolor{cyan!10}36.54 & \cellcolor{cyan!10}21.33 & \cellcolor{cyan!10}45.67 & \cellcolor{cyan!10}47.79 & \cellcolor{cyan!10}37.60 & \cellcolor{cyan!10}49.56 & \cellcolor{cyan!10}51.87 & \cellcolor{cyan!10}42.62 \\
\textsc{ARMOR} & \cellcolor{cyan!25}\textbf{77.81} & \cellcolor{cyan!25}\textbf{78.09} & \cellcolor{cyan!25}\textbf{75.59} & \cellcolor{cyan!25}\textbf{42.67} & \cellcolor{cyan!25}\textbf{45.62} & \cellcolor{cyan!25}\textbf{29.33} & \cellcolor{cyan!25}\textbf{66.93} & \cellcolor{cyan!25}\textbf{66.47} & \cellcolor{cyan!25}\textbf{58.67} & \cellcolor{cyan!25}\textbf{62.47} & \cellcolor{cyan!25}\textbf{63.39} & \cellcolor{cyan!25}\textbf{54.53} \\
\bottomrule
\end{tabular}
}}
\caption{Performance comparison of \textsc{ARMOR} with baselines. We report Correctness (Cor.), F1 score, and Exact Match (EM) in percentage (\%) using Qwen-3-1.7B and Llama-3.2-3B as backbones. The best results are highlighted in \textbf{bold}. Full results are provided in Appendix \ref{app_sec:add_exp}.}
\label{tab:main_result}
\end{table*}
\vspace{-0.3em}

\subsection{Main Results}
As demonstrated in our preliminary study (Appendix~\ref{app_sec:Retrieval Noise Distribution}), standard retrieval systems inevitably retrieve irrelevant or inconsequential documents, introducing \textit{natural noise} into the generation process.
Table~\ref{tab:main_result} presents the comparative performance of \textsc{ARMOR} and baselines under this realistic setting.
\textsc{ARMOR} achieves SOTA performance across all datasets and metrics, demonstrating remarkable consistency on both Qwen and Llama backbones.
Specifically, \textsc{ARMOR} surpasses the strongest baseline (\textsc{IRCoT}) by a substantial margin, achieving an average Correctness improvement of 38.8\% on Qwen-3-1.7B and 19.9\% on Llama-3.2-3B.
Standard \textsc{Vanilla-RAG} shows limited improvement over direct inference (\textsc{No-RAG}) and performs poorly on complex datasets.
This confirms that without explicit robustness mechanisms, generators are easily misled by the inherent noise in natural retrieval.
\textsc{IRCoT} remains competitive by interleaving reasoning and retrieval steps.
\textsc{CoN} struggles to maintain stability, suggesting that their sequential noise-filtering logic is prone to error propagation in open-domain settings.
\textsc{ARMOR}'s decoupled training objective effectively mitigates this issue.
Existing RL-based approaches (\textsc{RAG-RL}, \textsc{Search-R1}) exhibit suboptimal performance, highlighting the difficulty of optimizing RAG generators with sparse rewards under natural noise.

\begin{figure*}
    \centering
    \includegraphics[width=\textwidth]{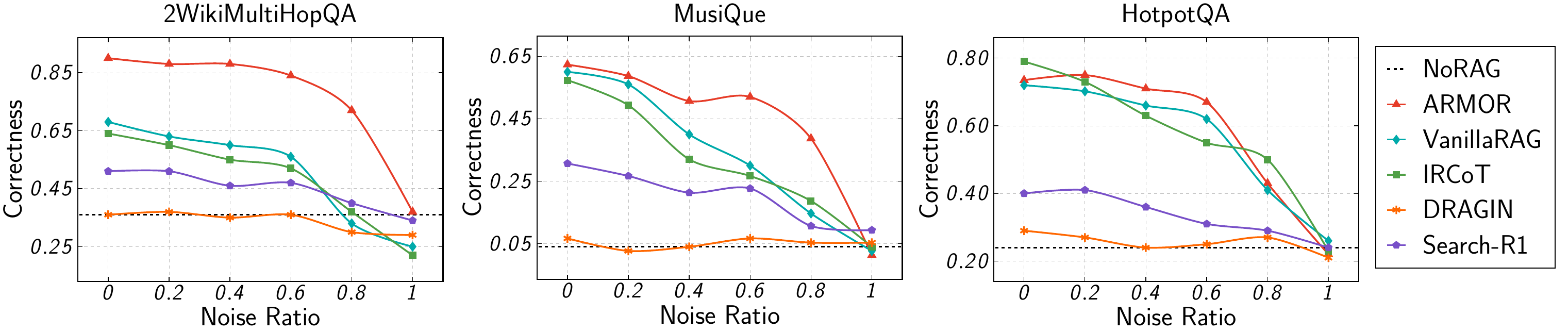}
    \caption{Robustness analysis under controlled noise ratios (from 0\% to 100\%) with fixed retrieval size of $K=5$.}
    \label{fig:Noise-Ratio}
\end{figure*}

We conducted a controlled experiment by manually injecting noise into the retrieval.
We fixed the number of retrieved documents to 5 and varied the noise ratio from 0\% to 100\%, as shown in Figure~\ref{fig:Noise-Ratio}.
Baseline methods exhibit a \textbf{precipitous performance decline} as the noise ratio increases.
\textsc{IRCoT} and \textsc{Search-R1} suffer significant losses, indicating they struggle to filter out distracting information effectively when valid evidence is scarce.
In contrast, \textsc{ARMOR} demonstrates exceptional stability, maintaining a high correctness score of 69\% even at an 80\% noise ratio.
\textsc{ARMOR} often outperforms the \textsc{Direct Answer} baseline or remains comparable at 100\% noise.
This indicates that \textsc{ARMOR} has effectively learned a \textit{rejection mechanism} via \textbf{Inter-Group Advantage Estimation}, allowing it to suppress the influence of irrelevant contexts rather than being hallucinated.

\subsection{Ablation Study}
We compare ARMOR with the following variants: 
(1) \textit{w/o CRL}: replaces the ALP-based scheduler with random sampling~(\textit{-random}), rigid sequential ordering~(\textit{-seq}), or outcome-supervised training~(\textit{-outcome sup}:~removing process rewards); 
(2) \textit{w/o Inter-Group Adv}: reverts global advantage normalization to standard intra-group normalization; 
(3) \textit{Base Setting}: removes both adaptive curriculum and inter-group estimation, reverting to standard outcome-supervised RL.
As shown in Table~\ref{tab:ablation_study}, ARMOR consistently outperforms all variant settings across all metrics.
Specifically, \textit{w/o CRL -seq} suffers a 47.51\% drop in average correctness, indicating that rigid task ordering leads to \textbf{catastrophic forgetting} among decoupled sub-tasks.
Similarly, \textit{w/o CRL -random} underperforms relative to ARMOR, confirming that the ALP-based scheduler is essential for dynamically balancing exploration and exploitation based on learning feedback.
Reverting to standard intra-group normalization (\textit{w/o inter-group adv.}) degrades performance by 14.58\%. 
This decline highlights that standard intra-group normalization \textbf{yields unstable signals in our decoupled setting}.
In contrast, ARMOR's batch-level advantage estimation provides robust supervision, effectively preventing mode collapse and stabilizing optimization.

\begin{table*}[hbtp]
\centering
\resizebox{\textwidth}{!}{
\scalebox{0.8}{
    \begin{tabular}{lcccccccccccc}
    \toprule
    \multirow{2}[1]{*}{\textbf{Method}} & \multicolumn{3}{c}{\textit{2WikiMultiHopQA}} & \multicolumn{3}{c}{\textit{MusiQue}} & \multicolumn{3}{c}{\textit{HotpotQA}} & \multicolumn{3}{c}{\textit{Avg.}} \\
    \cmidrule(lr){2-4} \cmidrule(lr){5-7} \cmidrule(lr){8-10} \cmidrule(lr){11-13}
    & \textbf{Cor.} & \textbf{F1} & \textbf{EM} & \textbf{Cor.} & \textbf{F1} & \textbf{EM} & \textbf{Cor.} & \textbf{F1} & \textbf{EM} & \textbf{Cor.} & \textbf{F1} & \textbf{EM} \\
    \midrule
    \rowcolor{gray!20} \multicolumn{13}{c}{\textbf{Qwen-3-1.7B}} \\
    \textbf{\textsc{ARMOR}} & \textbf{79.14} & \textbf{79.22} & \textbf{77.22} & \textbf{54.67} & \textbf{57.22} & \textbf{46.67} & \textbf{57.73} & \textbf{54.86} & \textbf{58.00} & \textbf{63.85} & \textbf{63.77} & \textbf{60.63} \\
    \textit{w/o CRL -random} & 67.66 & 67.97 & 66.48 & 39.33 & 40.59 & 28.67 & 37.27 & 35.45 & 27.33 & 48.09 & 48.00 & 40.83 \\
    \textit{w/o CRL -seq} & 52.22 & 52.69 & 50.89 & 23.65 & 26.50 & 15.20 & 24.67 & 21.76 & 11.27 & 33.51 & 33.65 & 25.79 \\
    \textit{w/o CRL -outcome sup.} & 60.22 & 60.45 & 58.59 & 32.33 & 32.37 & 23.00 & 30.80 & 27.72 & 18.73 & 41.12 & 40.18 & 33.44 \\
    \textit{w/o inter-group adv.} & 73.55 & 73.49 & 71.63 & 48.00 & 46.28 & 40.00 & 42.07 & 39.73 & 31.13 & 54.54 & 53.17 & 47.59 \\
    Base Setting & 71.63 & 72.04 & 70.30 & 44.33 & 45.79 & 36.67 & 40.47 & 38.29 & 30.07 & 52.14 & 52.04 & 45.68 \\
    \bottomrule
    \end{tabular}
    }}
    \caption{Ablation analysis on the effectiveness of the Adaptive Curriculum Learning~(CRL) strategy and Inter-Group Advantage Estimation~(inter-group adv). The best results are highlighted in \textbf{bold}.}
    \label{tab:ablation_study}
\end{table*}

\subsection{Analysis of Dynamic Learning Feedback}
\begin{figure}
    \centering
    \includegraphics[width=\linewidth]{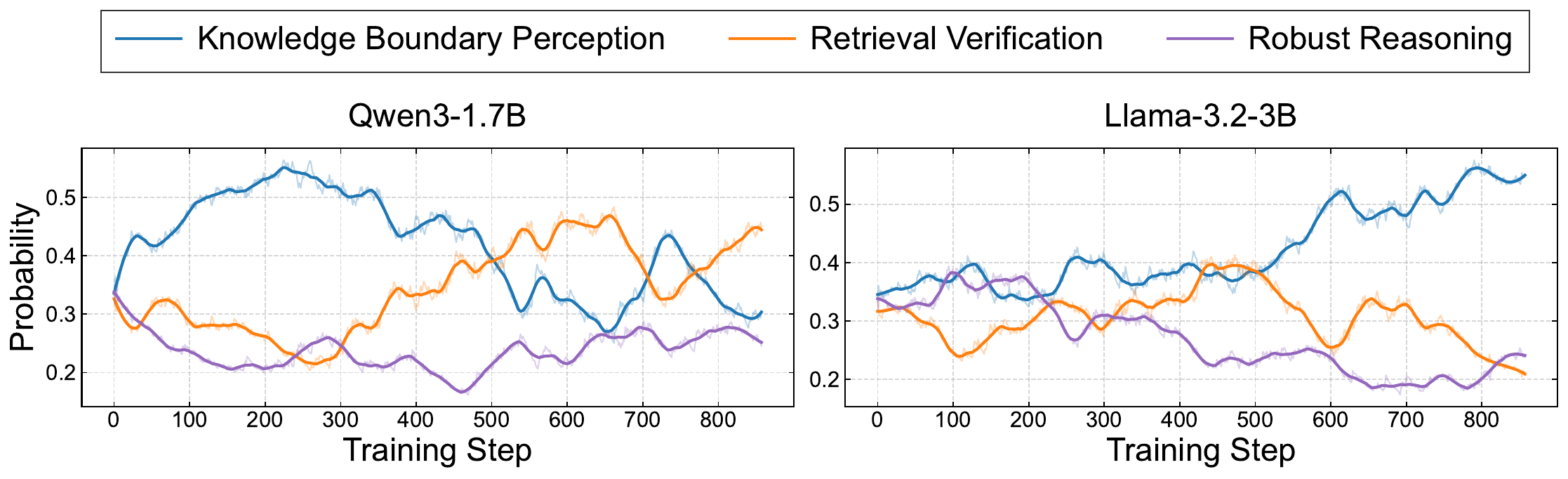}
    \caption{Task selection probabilities $p_n(\tau)$ during the training.}
    \vspace{-1em}
    \label{fig:curriculum_vis}
\end{figure}

We visualize the dynamic trends of the task selection probability $p_n(\tau)$ throughout the training process.
Figure~\ref{fig:curriculum_vis} illustrates these trends for both Llama-3.2-3B and Qwen-3-1.7B backbones.
Observing the training dynamics of Qwen-3-1.7B, the scheduler assigns high probability to the Knowledge Boundary Perception task in the early stages.
This indicates that the generator primarily focuses on distinguishing ``\textit{what it knows}" vs. ``\textit{what it needs to retrieve.}"
The scheduler shifts its focus toward Retrieval Verification around step 300.
This transition suggests that once the generator learns ``\textit{when to retrieve},'' it begins to focus on ``\textit{how to filter noise}'' from the retrieval.
Meanwhile, the attention on Robust Reasoning remains relatively stable, serving as a consistent regularization throughout the process.
The trends for Llama-3.2-3B exhibit a different pattern, highlighting the scheduler's adaptability to different generator architectures.
Unlike Qwen, Llama shows a continuous and increasing demand for Knowledge Boundary Perception. 
This suggests that for the Llama backbone, accurately determining the necessity of retrieval is a more challenging task.
These findings confirm that ARMOR does not impose a rigid \textit{``one-size-fits-all"} curriculum. 
Instead, it acts as a dynamic meta-learner, tailoring the training strategy to the learning trajectory of the underlying generator.

\subsection{Case Study}
We analyze a multi-hop query: ``\textit{When was Ole Rynning's father born?}'' (answer: ``\textit{May 30, 1778}''), detailed case is provided in Appendix~\ref{app_sec:case_study}.
\textsc{ARMOR} successfully decomposes the logic: identifying the father (``\textit{Jens Rynning}'') first, then retrieving his birth year. Crucially, ARMOR effectively filters out distractor documents describing the son's birth ``\textit{1809}'', leading to the correct answer.
In contrast, \textsc{Search-R1} exhibits chaotic reasoning, generating hallucinated contexts and failing to align the answer type.
\textsc{IRCoT} suffers from distraction failure, although it retrieves the correct evidence, it lacks explicit noise rejection and is misled by the high semantic overlap of the son's birth date.
This case validates that ARMOR's adaptive curriculum meta-learning effectively equips the generator with the ability to distinguish golden evidence from noise in complex reasoning scenarios.

\vspace{-0.3em}
\section{Conclusion}
In this work, we presented ARMOR, a novel framework designed to fortify RAG systems against the pervasive challenge of retrieval noise.
ARMOR decouples the training objective into three distinct sub-tasks: Knowledge Boundary Perception, Retrieval Verification, and Robust Reasoning.
To coordinate these tasks, ARMOR introduces an Adaptive Curriculum Meta-Learning strategy, where a \textit{teacher} scheduler dynamically prioritizes learning objectives based on the generator's Absolute Learning Progress.
Moreover, ARMOR addresses the optimization instability inherent in decoupled reinforcement learning by introducing Inter-Group Advantage Estimation, which mitigates mode collapse and ensures stable gradient optimization via global batch normalization.
Extensive experiments across three complex multi-hop reasoning datasets demonstrate that ARMOR exhibits exceptional resilience, maintaining high correctness even under extreme noise ratios.
This confirms that Adaptive Curriculum Scheduling and Inter-Group Advantage Estimation are essential for comprehensive RAG noise-robust reasoning.

\section{Limitations}
\textbf{Evaluation Scope.} 
Our experiments is confined to general-domain multi-hop QA datasets, including \textit{2WikiMultiHopQA}, \textit{MuSiQue} and \textit{HotpotQA}.
These datasets fail to capture the constraints and structural complexity of specialized domains. Consequently, the effectiveness of ARMOR in practical, high-stakes environments may be overestimated. Extending the evaluation to domains such as mathematics and healthcare is necessary to more rigorously assess its robustness and generalization.
\textbf{Language Scope.} 
Our experiments are limited to English-only corpora, which biases the evaluation toward high-resource language settings. Although the proposed approach is not inherently language-dependent, its performance under multilingual conditions—where retrieval quality, linguistic variation, and knowledge distribution differ significantly—remains underexplored. Systematically extending and validating ARMOR in cross-lingual settings is an important direction for future investigation.

\section*{Acknowledgments}
This work is partially supported by National Nature Science Foundation of China under No. 62476058. We thank the Big Data Computing Center of Southeast University for providing the facility support on the numerical calculations in this paper.

\bibliographystyle{named}
\bibliography{ijcai26}

\clearpage

\etocdepthtag.toc{atoc}

\etocsettagdepth{mtoc}{none} 
\etocsettagdepth{atoc}{subsection} 

\etocsettocstyle
    {\section*{Appendix Contents}} 
    {} 

\tableofcontents

\appendix

\section{Details of Dataset Reconstruction} \label{app_sec:Details of Dataset Reconstruction}
\begin{figure*}[h]
\centering
\setstretch{1.2}
\begin{tcolorbox}[
    width=0.9\textwidth,
    title=Structured Reasoning Chain,
    toptitle=0.3em,
    bottomtitle=0.3em
]

\textbf{Main Question:} When did the director of film The Argyle Secrets die?

\textbf{Main Answer:} April 16, 1995

\vspace{0.5em}

\textbf{Reasoning Node 1:}
\begin{enumerate}[leftmargin=2em, itemindent=0em, itemsep=-0.3em,label=\textbullet]
    \item \textbf{Intermediate Thought:} To find out when the director of The Argyle Secrets died, I first need to identify who directed the film.
    \item \textbf{Sub-Quention:} Who directed the film The Argyle Secrets?
    \item \textbf{Sub-Answer:} The Argyle Secrets was directed by Cy Endfield.
    \item \textbf{Retrieval Policy:} retrieve-then-answer
    \item \textbf{Golden Documents:}  
    \begin{enumerate}[itemsep=0em]
        \item The Arygle Secrets is a 1948 American mystery film directed by Cy Endfield and starring William Gargan and Marjorie Lord.
        \item In 1948, The Arygle Secrets, an American mystery film, was brought to life by director Cy Endfield, with notable performances from William Gargan and Marjorie Lord in leading roles.
        \item The Arygle Secrets, released in 1948, is an American mystery film directed by Cy Endfield, featuring performances by William Gargan and Marjorie Lord in the lead roles.
        \item ...
                    
    \end{enumerate}
\end{enumerate}

\vspace{0.5em}

\textbf{Reasoning Node 2:}
\begin{enumerate}[leftmargin=2em, itemindent=0em, itemsep=-0.3em,label=\textbullet]
    \item \textbf{Intermediate Thought:} Now that I know Cy Endfield directed The Argyle Secrets, I can look up his date of death to answer the main question.
    \item \textbf{Sub-Quention:} When did Cy Endfield die?
    \item \textbf{Sub-Answer:} The Argyle Secrets was directed by Cy Endfield.
    \item \textbf{Retrieval Policy:} direct-answer
    \item \textbf{Golden Documents:}  
    \begin{enumerate}[itemsep=0em]
        \item TCyril Raker Endfield (November 10, 1914 – April 16, 1995) was an American screenwriter, film director, theatre director, author, magician and inventor.
        \item Cyril Raker Endfield, born on November 10, 1914, and passing away on April 16, 1995, was a multifaceted American creative talent who excelled in various fields including screenwriting, filmmaking, theatre direction, authoring books, performing magic, and inventing.
        \item ...
    \end{enumerate}
\end{enumerate}

\end{tcolorbox} 
\caption{An example of the structured reasoning chain..}
\label{fig:reconstructed-dataset-demo}
\end{figure*}
We reconstruct \textit{2WikiMultiHopQA} and \textit{MuSiQue} datasets to train the RAG generator and evaluate its performance on multi-hop noise-robust reasoning.
As illustrated in Figure~\ref{fig:reconstructed-dataset-demo}, ARMOR requires training instances organized as structured reasoning chains, where each node in reasoning chains contains an intermediate thought, a sub-question, a retrieval policy, a sub-answer and a set of golden documents supporting the sub-answer.
However, the original versions of these datasets lack several node-level annotations.
To bridge this gap, we reconstruct each QA instance into a fully structured reasoning chain through a seven-stage pipeline:
(1) reasoning node construction, (2) sub-question construction, (3) sub-answer reorganization, (4) golden document augmentation, (5) retrieval policy annotation, and (6) intermediate thought synthesis (7) manual verification.

Several reconstruction stages rely on LLM-based generation.
To mitigate potential stylistic bias arising from over-reliance on a single generator, we employ an ensemble of advanced LLMs, including \texttt{GPT-5}~\cite{GPT-5}, \texttt{Gemini-2.5-Pro}~\cite{Gemini-2.5}, and \texttt{DeepSeek-R1}~\cite{Deepseek-R1}, and we prompt multiple models and randomly sample from the pool of valid outputs.
This strategy reduces the risk of introducing a discernible stylistic fingerprint into the synthetic data, thereby enhancing the ecological validity of the reconstructed datasets.

\subsection{Reasoning Node Construction}
This stage constructs a sequence of reasoning nodes.
For \textit{2WikiMultiHopQA}, a reasoning node is constructed from an evidence triple and its associated golden document.  
For \textit{MuSiQue}, the dataset provides a sequence of reasoning steps through an entity and a relation phrase. 
We construct each reasoning node through pairing the entity and relation phrase with their annotated sub-answer and supporting document.

\subsection{Sub-Question Construction}
For each reasoning node, we employ LLMs to formulate the sub-question that precisely reflects its reasoning objective.
In \textit{2WikiMultiHopQA}, sub-questions are generated based on the evidence triples and the main-question.
For the first node in reasoning chains, LLMs incorporate the main-question and evidence triple to generate the sub-question. 
For subsequent nodes, LLMs additionally incorporate previous sub-question-answer pairs to produce follow-up sub-questions that logically progress toward the final answer. The prompts used to generate the sub-queation are shown in Figure~\ref{fig:Prompt for generating later multi-hop sub-questions in 2WikiMultiHopQA} and Figure~\ref{fig:Prompt for generating the first-hop sub question in 2WikiMultiHopQA}, respectively.
In \textit{MuSiQue}, LLMs refine each node's entity and relation phrase into a complete sub-question, and the prompt is shown in Figure~\ref{fig:Prompt for reconstructing the sub question in MusiQue}.

\begin{figure*}[h]
\centering
\setstretch{1.2}
\begin{tcolorbox}[
    width=0.9\textwidth,
    title=2WikiMultiHopQA: First-Node Sub Question Generation,
    toptitle=0.3em,
    bottomtitle=0.3em
]

\textbf{Task Requirements:}

You are a reasoning assistant tasked with reconstructing the reasoning chain for a multi-hop question.
Your job is to generate \textbf{one natural language sub-question} based on the given inputs:

\begin{enumerate}[leftmargin=2em, itemindent=0em, itemsep=-0.3em,label=\textbullet]
    \item \textbf{Main Question}
    \item \textbf{Main Answer}
    \item \textbf{Evidence} (formatted as a triple: \emph{[subject, predicate, object]})
\end{enumerate}

The sub-question must:
\begin{enumerate}[leftmargin=2em, itemindent=0em, itemsep=-0.3em,label=\textbullet]
    \item Be answerable using \textbf{only} the given evidence triple.
    \item Serve as a logical intermediate step toward the main question.
    \item Explicitly mention the subject when appropriate, and use clear, concise language.
\end{enumerate}

Output \textbf{strictly} in the following \textbf{JSON} format. Do NOT add explanations or extra text.

\vspace{0.5em}
\textbf{Example Input:}

\emph{main question: Who wrote the novel that inspired the film Blade Runner?}\\
\emph{main answer: Philip K. Dick}\\
\emph{evidence: [novel Do Androids Dream of Electric Sheep?, basis, movie Blade Runner]}

\vspace{0.5em}
\textbf{Example Output:}

\emph{\{ ``sub-question": ``What novel is the basis for the movie Blade Runner?" \}}

\vspace{0.5em}
\textbf{Input:}

	\emph{main question: }\texttt{<}\emph{main question}\texttt{>}
    
    \emph{main answer: }\texttt{<}\emph{main answer}\texttt{>}
    
    \emph{evidence: }\texttt{<}\emph{evidence}\texttt{>}

\end{tcolorbox} 
\caption{Prompt for generating the first-hop sub question in 2WikiMultiHopQA.}
\label{fig:Prompt for generating the first-hop sub question in 2WikiMultiHopQA}
\end{figure*}
\begin{figure*}[htb]
\centering
\setstretch{1.2}
\begin{tcolorbox}[
    width=0.9\textwidth,
    title=2WikiMultiHopQA: Multi-Hop Sub Question (\nth{2}-Node Onward) Generation,
    toptitle=0.3em,
    bottomtitle=0.1em
]

\textbf{Task Requirements:}

You are a reasoning assistant tasked with reconstructing the reasoning chain for a multi-hop question.
Your job is to generate \textbf{one natural language sub-question} based on the given inputs:

\begin{enumerate}[leftmargin=2em, itemsep=-0.3em, itemindent=0em, label=\textbullet]
    \item \textbf{Main Question}
    \item \textbf{Main Answer}
    \item \textbf{Evidence} \\ (formatted as a triple: \emph{[subject, predicate, object]})
    \item \textbf{Previous Sub Question-Answer Pair}
    \\ \emph{(formatted as: [[sub-question-1, sub-answer-1], [sub-question-2, sub-answer-2], ...])}
    
\end{enumerate}

The sub-question must:
\begin{enumerate}[leftmargin=2em, itemsep=-0.3em, label=\textbullet]
    \item Be focused and answerable using \textbf{only} the given supporting fact.
    \item Together with prior sub Question-Answer steps, Represent a \textbf{logical intermediate step} toward answering the main question.
    \item Use \textbf{clear and concise language}, explicitly mentioning the **subject** when appropriate.
\end{enumerate}

Output \textbf{strictly} in the following \textbf{JSON} format. Do NOT add explanations or extra text.

\vspace{0.5em}
\textbf{Example Input:}

\emph{main question: Who wrote the novel that inspired the film Blade Runner?}

\emph{main answer: Philip K. Dick}

\emph{evidence: ["Do Androids Dream of Electric Sheep?", "author", "Philip K. Dick"]}

\emph{previous sub question-answer pair:}

\emph{[["What novel is the basis for the movie Blade Runner?", "Do Androids Dream of Electric Sheep?"]]}

\vspace{0.5em}
\textbf{Example Output:}

\emph{\{ ``sub-question": ``What is the name of the author of Do Androids Dream of Electric Sheep?" \}}

\vspace{0.5em}
\textbf{Input:}

	\emph{main question: }\texttt{<}\emph{main question}\texttt{>}
    
    \emph{main answer: }\texttt{<}\emph{main answer}\texttt{>}
    
    \emph{evidence: }\texttt{<}\emph{evidence}\texttt{>}
    
    \emph{previous sub question-answer pair: }\texttt{<}\emph{previous sub question-answer pair}\texttt{>}

\end{tcolorbox} 
\caption{Prompt for generating multi-hop sub-questions (second-hop onward) in 2WikiMultiHopQA.}
\label{fig:Prompt for generating later multi-hop sub-questions in 2WikiMultiHopQA}
\end{figure*}

\begin{figure}[H]
\centering
\setstretch{1.2}
\begin{tcolorbox}[width=0.45\textwidth, title=MusiQue: Sub Question Reconstruction,
toptitle={0.3em}, bottomtitle={0.1em}]

\textbf{Task Requirements:}

You will be given one question in the format:
\emph{Entity \texttt{>>} Relation}.
Your task is to convert each into a clear, natural-sounding English question. 
Use appropriate phrasing based on the relation (e.g., "distributed by", "owned by", etc.).
Do not add any extra information, explanations, or formatting. Please output only the rephrased question.

\vspace{0.5em}
\textbf{Example Input:}

\emph{The Game (1997 film) \texttt{>>} distributed by }

\vspace{0.5em}
\textbf{Example Output:}

\emph{Which company distributed \\ The Game (1997 film)?}

\vspace{0.5em}
\textbf{Input:}

	\emph{question: }\texttt{<}\emph{question}\texttt{>}

\end{tcolorbox} 
\caption{Prompt for reconstructing the sub question in MusiQue.}
\label{fig:Prompt for reconstructing the sub question in MusiQue}
\end{figure}

\subsection{Sub-Answer Reorganization}
This stage employs LLMs to reorganize the sub-answers in \textit{2WikiMultiHopQA} based on corresponding evidence triples. Specifically, sub-answers are normalized into concise phrases or sentences that directly address the corresponding sub-question. 
The prompt used for sub-answer reorganization is illustrated in Figure~\ref{fig:Prompt for reconstructing the sub answer in 2WikiMultiHopQA}.

\subsection{Golden Document Augmentation}
In this stage, we leverage LLMs to paraphrase the original golden document of each node into five distinct variants. This process ensures that the augmented documents retain the semantic equivalence of the original while introducing linguistic diversity. These generated documents are utilized to systematically manipulate the golden document ratio and noise ratio in noise-robust experiments. The prompt used for document augmentation is detailed in Figure~\ref{fig:Prompt for golden document augmentation}.

\subsection{Retrieval Policy Annotation}

We annotate each node by evaluating the RAG generator in retrieval-disabled mode. 
A node is labeled \textbf{retrieve-no} if the model generates a correct sub-answer based solely on its intrinsic knowledge and prior reasoning; otherwise, it is labeled \textbf{retrieve-yes}, indicating that external retrieval is required.
This model-specific annotation provides explicit supervision for training the generator to make adaptive retrieval policies during the warm-up stage.

\begin{figure}[H]
\centering
\setstretch{1.2}
\begin{tcolorbox}[width=0.45\textwidth, title=2WikiMultiHopQA: Sub Answer Reconstruction,
toptitle={0.3em}, bottomtitle={0.1em}]

\textbf{Task Requirements:}

You are an information conversion tool. Please convert the given triple \emph{[subject, predicate, object]} into a grammatically correct and natural-sounding English sentence.
Do not add any extra information, explanations, or formatting. Please output only the sentence.

\vspace{0.5em}
\textbf{Example Input:}

\emph{[novel Do Androids Dream of Electric Sheep?, basis, movie Blade Runner]}

\vspace{0.5em}
\textbf{Example Output:}

\emph{The novel Do Androids Dream of Electric Sheep is the basis for the movie Blade Runner.}

\vspace{0.5em}
\textbf{Input:}

	[\texttt{<}\emph{subject}\texttt{>}, \texttt{<}\emph{predicate}\texttt{>}, \texttt{<}\emph{object}\texttt{>}]

\end{tcolorbox} 
\caption{Prompt for reconstructing the sub answer in 2WikiMultiHopQA.}
\label{fig:Prompt for reconstructing the sub answer in 2WikiMultiHopQA}
\end{figure}

\subsection{Intermediate Thought Synthesis} 
This stage employ LLMs to synthesize an internal reasoning thought for each node, transforming the implicit logic into an explicit structured representation. Unlike simple concatenation of sub-questions and sub-answers, these states are generated by considering the global context, including the main question, the final answer, and the entire sequence of reasoning steps. 
Specifically, each synthesized thought elucidates the rationale behind its corresponding sub-question and justifies its necessity within the overall multi-hop chain, and the prompt is detailed in Figure~\ref{fig:Prompt for reconstructing the thinking process in 2WikiMultiHopQA and MusiQue}. These synthesized thoughts provide fine-grained supervision for the RAG generator to model internal reasoning processes during the warm-up stage.

\subsection{Manual Verification}

We conduct a manual verification stage to ensure the logical correctness of the reconstructed reasoning chains.
Specifically, annotators inspect each chain to identify and remove cases with logical inconsistencies, invalid reasoning transitions, or misaligned sub-questions and sub-answers.
This human-in-the-loop verification serves as a final quality control stage to guarantee the reliability of the reconstructed datasets.

\begin{figure*}[h]
\centering
\setstretch{1}
\begin{tcolorbox}[width=0.9\textwidth, title=Thinking Process Reconstruction,
toptitle={0.3em}, bottomtitle={0.1em}]

\textbf{Objective:}

You are a reasoning assistant designed to reconstruct the implicit thinking process in a multi-hop reasoning chain.
Given a main question, its final answer, and a sequence of reasoning steps, each includes a sub-question and sub-answer pair, your task is to generate a concise internal thought before each reasoning step.
Each internal thought should:
\begin{enumerate}[leftmargin=2em, itemindent=0em, label=\textbullet, itemsep=0em]
    \item Reflect the reasoning or purpose behind the corresponding sub-question.
    \item Explain why this step is necessary in the context of the overall reasoning process.
   	\item Logically connect to the previous steps when applicable.
   	\item Be succinct and focused on the rationale for that specific step.

\end{enumerate}
\vspace{0em}

\textbf{Task Requirements:}

\begin{enumerate}[leftmargin=2em, itemindent=0em, label=\textbullet, itemsep=0em]
    \item Input is provided in \textbf{JSON} format with fields: \emph{``main-question"}, \emph{``main-answer"}, and \emph{``reasoning-steps"} (a list of \emph{``sub-question"} and \emph{``sub-answer"}).

    \item Output must be a \textbf{JSON} format containing only the generated \emph{``thought"} strings, in the same order as the reasoning steps.

    \item Each \emph{``thought"} must:
    \begin{enumerate}[itemsep=0em]
        \item Explain the rationale behind the corresponding sub-question.
        \item Be concise and logically connected to the overall chain.
        \item Do \textbf{not} return the original input or duplicate any fields.
    \end{enumerate}
\end{enumerate}
\vspace{0em}
\textbf{Example Input:}

\{
\begin{enumerate}[label={}, leftmargin=2em, itemindent=0em, itemsep=0em]
    \item \emph{``main-question"}: \emph{``Who wrote the novel that inspired the film Blade Runner?",}

    \item \emph{``reasoning-steps":} \{
    \begin{enumerate}[label={}, leftmargin=2em, itemindent=0em, itemsep=0em]
        \item \emph{``step-1":} [
        \begin{enumerate}[label={}, leftmargin=2em, itemindent=0em, itemsep=0em]
            \item \emph{``What novel is the basis for the movie Blade Runner?",}
            \item \emph{``The novel Do Androids Dream of Electric Sheep? is the basis for the movie Blade Runner."}
        \end{enumerate}
        \item ]

        \item \emph{``step-2":} [
        \begin{enumerate}[label={}, leftmargin=2em, itemindent=0em]
            \item \emph{``What is the name of the author of Do Androids Dream of Electric Sheep?",}
            \item \emph{``The novel Do Androids Dream of Electric Sheep? was written by Philip K. Dick."}
        \end{enumerate}
        \item ],
    \end{enumerate}
    \item \}

    \item \emph{``main-answer": ``Philip K. Dick"}
\end{enumerate}
\}

\vspace{0.5em}
\textbf{Example Output:}

\{
\begin{enumerate}[label={}, leftmargin=2em, itemindent=0em, itemsep=0em]
    \item \emph{``step-1": ``To identify the author of the novel that inspired \emph{Blade Runner}, I first need to determine which novel served as the source material for the film.",}
    \item \emph{``step-2": ``Now that I know the novel is \emph{Do Androids Dream of Electric Sheep?}, I need to find its author to answer the main question."}
\end{enumerate}
\}

\vspace{0.5em}

\textbf{Input:}

\{ 

\begin{enumerate}[label={}, leftmargin=2em, itemindent=0em, itemsep=0em]
    \item \emph{``main-question":} \texttt{<}\emph{main question}\texttt{>},
    \item \emph{``reasoning-steps":} \texttt{<}\emph{reasoning steps}\texttt{>},
    \item \emph{``main answer":} \texttt{<}\emph{main answer}\texttt{>}
\end{enumerate}

\}

\end{tcolorbox} 
\caption{Prompt for reconstructing the thinking process in 2WikiMultiHopQA and MusiQue.}
\label{fig:Prompt for reconstructing the thinking process in 2WikiMultiHopQA and MusiQue}
\end{figure*}

\begin{figure*}[h]
\centering
\setstretch{1.2}
\begin{tcolorbox}[width=0.9\textwidth, title=Prompt for Golden Document Augmentation,
toptitle={0.3em}, bottomtitle={0.1em}]

\textbf{Task Requirements:}

	Please generate \texttt{<}\emph{\#document number}\texttt{>} documents based on the input document below. 
	Each generated document should have the same meaning as the input document but be rephrased differently.
	Return the results in JSON format where the keys are integers from 1 to \texttt{<}\emph{\#document number}\texttt{>}, and the values are the rephrased documents.
\vspace{0.5em}

\textbf{Document:}

	\texttt{<}\emph{document}\texttt{>}

\vspace{0.5em}

\textbf{Example Input:}

	\emph{The Radio station currently plays a mix of Hindi and Regional music.}

\vspace{0.5em}
\textbf{Example Output:}

	\{

	\hspace{2em}\emph{1: "The Radio station is broadcasting a combination of Hindi and Regional music.",}

	\hspace{2em}\emph{2: "A mix of Hindi and Regional music is being played on the Radio station.",}

	\hspace{2em}\emph{3: "The Radio station features both Hindi and Regional music in its current playlist."}

	\}

\end{tcolorbox} 
\caption{Prompt for golden document augmentation.}
\label{fig:Prompt for golden document augmentation}
\end{figure*}

\section{Additional Details of ARMOR}
\subsection{Pseudo-Code of ARMOR}

\begin{algorithm}[tb]
    \small
    \caption{Adaptive Task Scheduler}
    \label{alg:adaptive_scheduler}
    \textbf{Input}: Task set $\mathcal{T}$, total training rounds $T$, training dataset $\mathcal{D}$\\
    \textbf{Hyperpara}: Exploration coefficient $\gamma$,  forgetting factor $\alpha$\\
    \textbf{Output}: Optimized RAG generator $\pi_{\theta}$
    
    \begin{algorithmic}[1] 
        \STATE Initialize learning feedback weight 
        \\ $w_{1}(\tau) \gets \frac{1}{\mathcal{T}}, \forall \tau \in \mathcal{T}$
        \STATE Initialize reward history 
        \\ $\mathcal{H}(\tau) \gets \emptyset, \forall \tau \in \mathcal{T}$
        \FOR{$n = 1$ \TO $T$}
            \STATE $W_{sum} \gets \sum_{\tau' \in \mathcal{T}} w_t(\tau')$
            \FOR{each task $\tau \in \mathcal{T}$}
                \STATE $ p_n(\tau) \gets (1 - \gamma) \frac{w_n(\tau)}{W_{sum}} + \frac{\gamma}{|\mathcal{T}|}$
            \ENDFOR
            
            \STATE Sample a task $\tau_s \sim p_n(\tau)$
            \STATE Select new batch of questions $\mathcal{Q}$ from $\mathcal{D}$ 
            \STATE Update $\theta_{G}$ via inter group advantage estimation \\ $\pi_{\theta} \gets \text{Optimizer}(\pi_{\theta}, \tau_s, \mathcal{Q})$
            \STATE Observe task-level reward $r_n(\tau_s)$
            
            \IF{$\mathcal{H}(\tau_s) \neq \emptyset$}
                \STATE $r_{\text{old}}(\tau_s) \gets \max \{ r \in \mathcal{H}(\tau_s) \}$
                \STATE $ALP(\tau_s) \gets |r_n(\tau_s) - r_{\text{old}}(\tau_s)|$
            \ELSE
                \STATE $ALP(\tau_s) \gets 0$
            \ENDIF
            \STATE $\mathcal{H}(\tau_s) \gets \mathcal{H}(\tau_s) \cup \{ r_n(\tau_s) \}$
            
            \FOR{each task $\tau \in \mathcal{T}$}
                \IF{$\tau = \tau_s$}
                    \STATE $\hat{r}_n(\tau) \gets ALP(\tau_s) / p_t(\tau_s)$
                \ELSE
                    \STATE $\hat{r}_n(\tau) \gets 0$
                \ENDIF
                \STATE $w_{n+1}(\tau) \gets w_n(\tau) \cdot \exp\left( \frac{\gamma \hat{r}_n(\tau)}{|\mathcal{T}|} \right) + \frac{e \cdot \alpha}{|\mathcal{T}|} \cdot W_{\text{sum}}$
            \ENDFOR
        \ENDFOR
        \STATE \textbf{return} RAG generator $\pi_{\theta}$
    \end{algorithmic}
\end{algorithm}

In this appendix, we provide the pseudo-code for the key components of the ARMOR framework: \textbf{Adaptive Curriculum Scheduler} and \textbf{Inter-Group Advantage Estimation}.
The Adaptive Curriculum Scheduler (Algorithm~\ref{alg:adaptive_scheduler}) dynamically select tasks for optimization during each training round. 
It begins by initializing the task weights and reward history. The scheduler computes task selection probabilities based on the current learning progress, balancing exploration and exploitation. 
The learning progress is assessed using ALP, and task weights are updated after each round to reflect learning progress while retaining historical experience with a forgetting factor.

The Inter-Group Advantage Estimation (Algorithm~\ref{alg:Iner Group Advantage Estimation}) normalizes task-level and format-level rewards across the entire batch. Multiple rollouts are generated for each question, and rewards are assigned for each task span. These rewards are then normalized to compute task-level and format-level advantages. The policy model is updated using a clipped objective that incorporates the computed advantages, ensuring stable optimization despite task heterogeneity.
Together, these two components enable ARMOR to provide an adaptive, stable, and efficient learning framework for noise-robust reasoning.

\begin{algorithm}[htbp]
    \small
    \caption{Iner Group Advantage Estimation}
    \label{alg:Iner Group Advantage Estimation}
    \textbf{Input}: Questions $\mathcal{Q}$, selected task $\tau_s$,  \\
    policy model $\pi_\theta$, reward model $\pi_{\text{reward}}$,  
     \\
    \textbf{Hyperpara}: Question number $m$, rollout number $l$, \\ format weight $\omega$,  
    clipping coefficient $\varepsilon$, current round $n$ \\
    \textbf{Output}: Updated policy parameters $\theta$

    \begin{algorithmic}[1]
        \STATE Initialize rollout set $\mathcal{O}_{\text{train}} \gets \emptyset$
        \STATE Sync $\pi_{\theta_{\text{old}}} \gets \pi_{\theta}$
        \FOR{$i = 1$ \TO $m$}
            \FOR{$j = 1$ \TO $l$}
                \STATE Sample rollout sequence $o_{i,j} \sim \pi_{\theta_{\text{old}}}(\cdot \mid q_i)$
                \STATE $\mathcal{O}_{\text{train}} \gets \mathcal{O}_{\text{train}} \cup \{o_{i,j}\}$
            \ENDFOR
        \ENDFOR

        \STATE Initialize $R_{\text{task}} \gets \emptyset$
        \FOR{each $o_{i,j} \in \mathcal{O}_{\text{train}}$}
            \STATE Identify $K_{i,j}$ task-spans in $o_{i,j}$ for task $\tau_s$
            \STATE Compute rewards $\mathbf{r}_{i,j} = \{ r_{i,j}^{(1)}, \dots, r_{i,j}^{(K_{i,j})} \}$ via $\pi_{\text{reward}}$
            \STATE $R_{\text{task}} \gets R_{\text{task}} \cup \{ \mathbf{r}_{i,j} \}$
        \ENDFOR

        \STATE Initialize $R_{\text{fmt}} \gets \emptyset$
        \FOR{each $o_{i,j} \in \mathcal{O}_{\text{train}}$}
            \STATE Compute format reward $r_{i,j}^{\text{fmt}}$
            \STATE $R_{\text{fmt}} \gets R_{\text{fmt}} \cup \{ r_{i,j}^{\text{fmt}} \}$
        \ENDFOR

        
        \FOR{each $o_{i,j} \in \mathcal{O}_{\text{train}}$}
            \FOR{$k = 1$ \TO $K_{i,j}$}
                \STATE $A_{i,j}^{\text{task}(k)} \gets \dfrac{r_{i,j}^{(k)} - \text{mean}(R_{\text{task}})}{\text{std}(R_{\text{task}})}$
            \ENDFOR
            \STATE $A_{i,j}^{\text{fmt}} \gets \dfrac{r_{i,j}^{\text{fmt}} - \text{mean}(R_{\text{fmt}})}{\text{std}(R_{\text{fmt}})}$
            
            \FOR{$t = 1$ \TO $|o_{i,j}|$}
                \STATE $ A_{i,j}^{(t)} = \sum_{k=1}^{K_{i,j}} \mathbf{1}[t \in \text{tokens}(i, j, k)] \cdot A_{i,j}^{\text{task}(k)}+\omega A_{i,j}^{\text{fmt}} $
            \ENDFOR
        \ENDFOR

        \STATE Initialize loss $\mathcal{L} \gets 0$
        \FOR{$i = 1$ \TO $m$}
            \STATE $L_i \gets 0$, $N_i \gets 0$
            \FOR{$j = 1$ \TO $n$}
                \FOR{$t = 1$ \TO $|o_{i,j}|$}
                    \STATE Compute likelihood ratio:
                    \\ $\rho_{i,j}^{(t)} \gets \dfrac{\pi_{\theta}\left(o_{i,j}^{(t)} \mid q_i, o_{i,j}^{(<t)}\right)}{\pi_{\theta_{\text{old}}}\left(o_{i,j}^{(t)} \mid q_i, o_{i,j}^{(<t)}\right)}$
                    \STATE Clipped objective:
                    \\ $c_{i,j}^{(t)} \gets \min\left( \rho_{i,j}^{(t)} A_{i,j}^{(t)},\; \text{clip}(\rho_{i,j}^{(t)}, 1-\varepsilon, 1+\varepsilon) \cdot A_{i,j}^{(t)} \right)$
                    \STATE $L_i \gets L_i + c_{i,j}^{(t)}$
                    \STATE $N_i \gets N_i + 1$
                \ENDFOR
            \ENDFOR
            \STATE $\mathcal{L} \gets \mathcal{L} + \dfrac{L_i}{N_i}$
        \ENDFOR
        \STATE Update $\theta$ by maximizing $\mathcal{L}$

        \STATE Rollout on validation set using $\pi_{\theta}$
        \STATE Compute $r_n(\tau_s) \gets \text{mean}(R_{\text{task}}^{\text{val}})$
        \STATE \textbf{return} policy mode $\pi_\theta, r_n(\tau_s)$
    \end{algorithmic}
\end{algorithm}

\subsection{Proof of Unbiased Reward Estimation}\label{app_sec:Proof of Unbiased Reward Estimation}

In our adaptive curriculum meta-learning framework, The scheduler samples a task $\tau_s \sim p_n(\cdot)$ at the start of round $n$.
Directly using the observed $ALP(\tau_s)$ to update task weights carries the risk of  underestimating rarely selected tasks and overestimating frequently selected tasks. To ensure fair credit assignment across all tasks, we apply importance weighting to construct an unbiased estimation for $ALP(\tau)$.
The estimated reward is defined as:

\begin{equation}
    \hat{r}_n(\tau) = \frac{ALP(\tau)}{p_n(\tau)}.
\end{equation}
\begin{proof}
\( \hat{r}_n(\tau) \) is an unbiased estimation of \( ALP(\tau) \).\\
\begin{equation}
\begin{aligned}
\mathbb{E}[\hat{r}_n(\tau)] 
&= \mathbb{E}\!\left[\frac{ALP(\tau)}{p_n(\tau)} \right] \\
&= \frac{ALP(\tau)}{p_n(\tau)} \cdot p_n(\tau) \\
&= ALP(\tau).
\end{aligned}
\end{equation}
Hence, $\hat{r}_n(\tau)$ is an unbiased estimation for $ALP(\tau)$.
\end{proof}

\section{Additional Experimental Setup} \label{app_sec:additional_exp_setup}

\subsection{Baseline Configurations}

To ensure a fair and consistent comparison, we employ unified LLM generators and evaluation metrics across all experiments, while adhering to the originally recommended hyperparameter settings for each baseline.
\paragraph{Baseline Settings.}
For \textsc{DRAGIN}, we follow the original configuration by setting the number of few-shot examples to six and the hallucination threshold to 1.0, thereby effectively evaluating the output confidence.
Regarding \textsc{IRCoT}, we configure the maximum inference hops to four, with a retrieval limit of 15 documents per query. The generation of tokens is limited to a maximum of 248 tokens per hop.
For \textsc{ARMOR}, we mask all \verb|<doc>| tags and  their associated documents during training to prevent information leakage. We employ LoRA with a rank of $r=8$, a scaling factor of $\alpha=16$, and disable dropout.
For \textsc{Search-R1} and \textsc{RAG-RL}, we standardize the maximum sequence length to 1,500 tokens and perform six rollouts during the training process, maintaining the same LoRA configuration as \textsc{ARMOR}. During the rollout process, we adopt a sampling temperature of 0.85, complemented by $\text{top-}p=0.95$ and $\text{top-}k=50$. The retrieval module is configured to retain the top three documents for each query. Specifically for \textsc{RAG-RL}, we implement the min-max curriculum strategy, which is empirically identified as the optimal setting in our preliminary study.
In addition, it is crucial to note that \textsc{Search-R1} and \textsc{RAG-RL} are initialized directly from the original instruct-tuned model, rather than the model processed through our warm-up stage. This design choice is motivated by the fact that our warm-up stage is specifically engineered to instill particular reasoning formats. Since \textsc{Search-R1} and \textsc{RAG-RL} adhere to their own distinct structural paradigms, transferring the format biases from our warm-up model would create conflicting learning objectives, resulting in training collapse.

\paragraph{Inference and Environment.}
During the inference phase, all baselines and ARMOR utilize \texttt{e5-base-v2} as the unified retriever, retrieving up to three documents per iteration from the consistent corpus. Sampling hyperparameters (\ie temperature, $\text{top-}k$, and $\text{top-}p$) follow the default settings of the original instruct-tuned models. All experiments are conducted on four NVIDIA A100 GPUs. To ensure statistical reliability, reported results are averaged over three independent runs (using random seed 42 for initialization).

\subsection{Retrieval Noise Distribution} \label{app_sec:Retrieval Noise Distribution}

Standard RAG systems inevitably retrieve irrelevant or distracting documents, which introduces \textit{natural noise} into the generation process. To quantify this phenomenon, we analyze the composition of retrieved documents across \textit{2Wiki}, \textit{MuSiQue}, and \textit{HotpotQA} using \texttt{e5-base-v2} as the retriever with retreieve $k=3$. Specifically, for both the original multi-hop questions and their decomposed sub-questions, we categorize each retrieved document into three types: \textbf{Golden} (directly supports the answer), \textbf{Inconsequential} (semantically related but not essential), and \textbf{Irrelevant} (unrelated to the question).

\begin{table}[H]
    \centering
    \begin{tabularx}{\linewidth}{l>{\centering\arraybackslash}X>{\centering\arraybackslash}X>{\centering\arraybackslash}X}
        \toprule
        \textbf{Datasets} &
        \small\textbf{Golden} &
        \small\textbf{Inconsequential} &
        \small\textbf{Irrelevant} \\
        \midrule
        \rowcolor{gray!20}\multicolumn{4}{c}{\small\textbf{Multi-Hop Question}}\\
        \textit{2Wiki}    & 21.87   & 30.03 & 48.10 \\
        \textit{MuSiQue}  & 12.92   & 35.65 & 51.43 \\
        \textit{HotpotQA}  & 17.14   & 33.89 & 48.97 \\
        \midrule
        \rowcolor{gray!20}\multicolumn{4}{c}{\small\textbf{Decomposed Sub-Question}}\\
        \textit{2Wiki}    & 25.06   & 29.17 & 45.77 \\
        \textit{MuSiQue}  & 16.67   & 34.55 & 48.78 \\
        \textit{HotpotQA} & 13.33   & 33.67 & 53.00 \\
        \bottomrule
    \end{tabularx}
    \caption{Distribution of retrieved document types for multi-hop questions and their decomposed sub-questions across three datasets.}
    \label{tab:Experiment dataset statistics}
\end{table}

The results in Table~\ref{tab:Experiment dataset statistics} confirm that \textbf{retrieval noise is inherent in standard RAG pipelines}: a significant fraction of retrieved passages are either inconsequential or irrelevant. Multi-hop questions generally exhibit higher noise ratios than their decomposed sub-questions, underscoring the greater challenge posed by compositional questions to RAG retrievers. 
Furthermore, the retrieval noise ratios exhibit a progressive increase across three datasets: \textit{2WikiMultiHopQA} shows the lowest noise ratio while \textit{MuSiQue} and \textit{HotpotQA} exhibit higher noise ratios, correlating with growing reasoning complexity or entity diversity.
These findings highlight the necessity of noise-robust reasoning for multi-hop QA.

\subsection{Experimental Prompts}

This section provides an overview of the prompts employed during the inference stage of \textsc{ARMOR}.
Figure~\ref{fig:Prompt for ARMOR inference} illustrates the inference prompt used in \textsc{ARMOR}. When the model generates the token sequence \texttt{<search>True</search>}, the retriever is activated to fetch relevant documents based on the preceding \texttt{<sub-question>}. Each retrieved document is then inserted sequentially into the context; after each insertion, the RAG generator produces a corresponding \texttt{<doc-type>} annotation.
In addition, Figure~\ref{fig:Prompt for direct answer} shows the prompt template used to generate answers without retrieval. This variant is not part of the main inference pipeline but is instead utilized during the reinforcement learning phase to train the Knowledge Boundary Perception task, where the model learns to recognize when its internal knowledge suffices to answer a question.

\begin{figure}[H]
\centering
\setstretch{1.2}
\begin{tcolorbox}[width=0.45\textwidth, title=Prompt for Direct Answer,
toptitle={0.3em}, bottomtitle={0.1em}]

\textbf{Task Requirements:}

Please provide a direct and concise answer to the question. 
Do not add explanations or extra text.

\vspace{0.5em}
\textbf{Input:}

	\emph{question: }\texttt{<}\emph{question}\texttt{>}

\end{tcolorbox} 
\caption{Prompt for direct answer.}
\label{fig:Prompt for direct answer}
\end{figure}

\begin{figure*}[h]
\centering
\setstretch{1.2}
\begin{tcolorbox}[width=0.9\textwidth, title=Inference Prompt of ARMOR,
toptitle={0.3em}, bottomtitle={0.1em}]

\textbf{Task Requirements:}

You are a reasoning assistant. Answer \texttt{<}main-question\texttt{>} by performing explicit, step-by-step reasoning.
Follow this structure \textbf{exactly}, and do \textbf{not} include any text outside these tags:

\begin{enumerate}[leftmargin=2em, itemsep=-0.3em, itemindent=0em, label=\arabic*.]

\item  \texttt{<}think\texttt{>} ... \texttt{<}/think\texttt{>} — Describe your internal reasoning process for the current step.

\item  \texttt{<}sub-question\texttt{>} ... \texttt{<}/sub-question\texttt{>} — State the next sub-question needed to continue reasoning.

\item \texttt{<}search\texttt{>} True \texttt{<}/search\texttt{>} or \texttt{<}search\texttt{>} False \texttt{<}/search\texttt{>} — Indicate whether external retrieval is required for this sub-question.
\item  \texttt{<}doc\texttt{>} ... \texttt{<}/doc\texttt{>} — Include the full content of any retrieved documents. Skip if none.
\item  \texttt{<}doc-type\texttt{>}golden\texttt{<}/doc-type\texttt{>} or \texttt{<}doc-type\texttt{>}inconsequential\texttt{<}/doc-type\texttt{>} \texttt{<}doc-type\texttt{>}irrelevant\texttt{<}/doc-type\texttt{>} — Classify each retrieved document.
\item  \texttt{<}sub-answer\texttt{>} ... \texttt{<}/sub-answer\texttt{>} — Provide the answer to the current sub-question.

\end{enumerate}

Repeat steps 1–6 as needed for multiple reasoning steps.
Finally, use \texttt{<}main-answer\texttt{>} ... \texttt{<}/main-answer\texttt{>} to provide the final answer to the main question.

\vspace{0.5em}
\textbf{Example:}

\texttt{<}main-question\texttt{>}Who is younger, Marvin E. Miller Sr. or John Hawkins (Author)?\texttt{<}/main-question\texttt{>}

\texttt{<}think\texttt{>}Find the birth years of both people.\texttt{<}/think\texttt{>}

\texttt{<}sub-question\texttt{>}When was Marvin E. Miller Sr. born?\texttt{<}/sub-question\texttt{>}

\texttt{<}search\texttt{>}True\texttt{<}/search\texttt{>} 

\texttt{<}doc\texttt{>}Marvin E. Miller Sr. was born in 1917.\texttt{<}/doc\texttt{>} \texttt{<}doc-type\texttt{>}useful\texttt{<}/doc-type\texttt{>} 

\texttt{<}sub-answer\texttt{>}1917\texttt{<}/sub-answer\texttt{>}
\texttt{<}think\texttt{>}Find John Hawkins' birth year.\texttt{<}/think\texttt{>}

\texttt{<}sub-question\texttt{>}When was John Hawkins (Author) born?\texttt{<}/sub-question\texttt{>}

\texttt{<}search\texttt{>}False\texttt{<}/search\texttt{>} 

\texttt{<}sub-answer\texttt{>}1960\texttt{<}/sub-answer\texttt{>}
\texttt{<}think\texttt{>}Compare birth years; the later year is younger.\texttt{<}/think\texttt{>}

\texttt{<}main-answer\texttt{>}John Hawkins (Author) is younger.\texttt{<}/main-answer\texttt{>}

\vspace{0.5em}
\textbf{Input:}

	\texttt{<}\emph{main-question (tag)}\texttt{>} \texttt{<}\emph{main question}\texttt{>} \texttt{<}\emph{/main-question (tag)}\texttt{>}

\end{tcolorbox} 
\caption{Inference prompt of ARMOR.}
\label{fig:Prompt for ARMOR inference}
\end{figure*}

\begin{table*}[hbtp]
\centering
\resizebox{\textwidth}{!}{
\begin{tabular}{lccccccccccccc}
\toprule
\multirow{2}[1]{*}{\textbf{Method}} & \multicolumn{3}{c}{\textit{2WikiMultiHopQA}} & \multicolumn{3}{c}{\textit{MusiQue}} & \multicolumn{3}{c}{\textit{HotpotQA}} & \multicolumn{3}{c}{\textit{Avg.}} \\
\cmidrule(lr){2-4} \cmidrule(lr){5-7} \cmidrule(lr){8-10} \cmidrule(lr){11-13}
& \textbf{Cor.} & \textbf{F1} & \textbf{EM} & \textbf{Cor.} & \textbf{F1} & \textbf{EM} & \textbf{Cor.} & \textbf{F1} & \textbf{EM} & \textbf{Cor.} & \textbf{F1} & \textbf{EM} \\
\midrule

\rowcolor{gray!20}\multicolumn{13}{c}{\textbf{Qwen-3-0.6B}} \\
\textsc{No-RAG} & 37.43 & 35.55 & 32.99 & 3.73 & 2.58 & 0.00 & 20.27 & 16.26 & 10.47 & 20.48 & 18.13 & 14.49 \\
\textsc{Vanilla-RAG}           & 34.47 & 28.16 & 21.15 & 14.67 & 11.63 & 2.67 & 34.87 & 28.24 & 21.27 & 28.00 & 22.68 & 15.03 \\
\textsc{CoT}           & 34.17 & 31.29 & 28.55 & 2.67 & 5.99 & 0.00 & 19.67 & 18.16 & 14.33 & 18.84 & 18.48 & 14.29 \\
\textsc{CoN}           & 56.15 & 19.33 & 9.91 & 19.73 & 15.98 & 8.27 & 46.13 & 32.50 & 26.53 & 40.67 & 22.60 & 14.90 \\
\textsc{IRCoT}         & 36.69 & 31.06 & 23.52 & 13.33 & 12.44 & 3.73 & 42.47 & 35.57 & 26.87 & 30.83 & 26.36 & 18.04 \\
\textsc{DRAGIN}        & 32.40 & 33.10 & 30.47 & 1.33 & 3.76 & 0.00 & 19.60 & 18.97 & 16.53 & 17.78 & 18.61 & 15.67 \\
\textsc{SKR}           & 36.83 & 33.53 & 29.29 & 12.00 & 4.72 & 0.00 & 24.87 & 23.03 & 16.60 & 24.57 & 20.43 & 15.30 \\
\textsc{RAG-RL}        & 37.58 & 31.24 & 25.49 & 18.73 & 15.82 & 3.47 & 36.97 & 32.22 & 25.67 & 31.09 & 26.43 & 18.21 \\
\textsc{Search-R1}     & 24.18 & 17.35 & 11.55 & 8.67 & 5.89 & 1.67 & 26.13 & 19.35 & 12.47 & 19.66 & 14.20 & 8.56 \\
\textsc{ARMOR}~(warm-up stage)  & \cellcolor{cyan!10}62.57 & \cellcolor{cyan!10}62.49 & \cellcolor{cyan!10}59.32 & \cellcolor{cyan!10}28.00 & \cellcolor{cyan!10}28.97 & \cellcolor{cyan!10}21.33 & \cellcolor{cyan!10}32.27 & \cellcolor{cyan!10}31.03 & \cellcolor{cyan!10}33.60 & \cellcolor{cyan!10}40.95 & \cellcolor{cyan!10}40.83 & \cellcolor{cyan!10}38.08 \\
\textsc{ARMOR}     & \cellcolor{cyan!25}\textbf{72.34} & \cellcolor{cyan!25}\textbf{73.31} & \cellcolor{cyan!25}\textbf{70.56} & \cellcolor{cyan!25}\textbf{49.33} & \cellcolor{cyan!25}\textbf{47.40} & \cellcolor{cyan!25}\textbf{34.67} & \cellcolor{cyan!25}\textbf{50.00} & \cellcolor{cyan!25}\textbf{49.43} & \cellcolor{cyan!25}\textbf{41.13} & \cellcolor{cyan!25}\textbf{57.22} & \cellcolor{cyan!25}\textbf{56.71} & \cellcolor{cyan!25}\textbf{48.79} \\
\midrule

\rowcolor{gray!20}\multicolumn{13}{c}{\textbf{Qwen-3-4B}} \\
\textsc{No-RAG} & 32.84 & 28.41 & 23.67 & 10.67 & 11.08 & 2.67 & 30.07 & 25.64 & 20.13 & 24.53 & 21.71 & 15.49 \\
\textsc{Vanilla-RAG}           & 66.12 & 63.75 & 57.99 & 24.00 & 21.13 & 16.00 & 58.67 & 56.64 & 50.13 & 49.60 & 47.17 & 41.37 \\
\textsc{CoT}           & 44.97 & 28.53 & 23.52 & 12.00 & 9.16 & 2.67 & 33.27 & 30.88 & 25.27 & 30.08 & 22.86 & 17.15 \\
\textsc{CoN}           & 65.83 & 24.19 & 17.16 & 36.00 & 25.64 & 16.00 & 59.13 & 45.12 & 38.57 & 53.65 & 31.65 & 23.91 \\
\textsc{IRCoT}         & 67.31 & 62.72 & 55.03 & 44.00 & 34.35 & 24.00 & 63.93 & 57.60 & 48.33 & 58.41 & 51.56 & 42.45 \\
\textsc{DRAGIN}        & 45.19 & 46.50 & 43.64 & 8.00 & 9.23 & 4.00 & 32.87 & 32.84 & 29.73 & 28.69 & 29.52 & 25.79 \\
\textsc{SKR}           & 36.24 & 25.44 & 18.05 & 16.00 & 8.56 & 1.33 & 30.00 & 22.86 & 16.60 & 27.41 & 18.95 & 11.99 \\
\textsc{RAG-RL}        & 69.26 & 70.15 & 68.79 & 25.67 & 22.26 & 18.00 & 61.27 & 59.64 & 59.40 & 52.07 & 50.68 & 48.73 \\
\textsc{Search-R1}     & 59.56 & 56.90 & 51.74 & 21.67 & 18.22 & 13.13 & 50.67 & 48.75 & 47.33 & 43.97 & 41.29 & 37.40 \\
\textsc{ARMOR}~(warm-up stage)  & \cellcolor{cyan!10}70.12 & \cellcolor{cyan!10}70.85 & \cellcolor{cyan!10}69.38 & \cellcolor{cyan!10}24.00 & \cellcolor{cyan!10}24.29 & \cellcolor{cyan!10}14.67 & \cellcolor{cyan!10}51.07 & \cellcolor{cyan!10}50.38 & \cellcolor{cyan!10}44.20 & \cellcolor{cyan!10}48.40 & \cellcolor{cyan!10}48.51 & \cellcolor{cyan!10}42.75 \\
\textsc{ARMOR}     & \cellcolor{cyan!25}\textbf{77.96} & \cellcolor{cyan!25}\textbf{77.44} & \cellcolor{cyan!25}\textbf{75.00} & \cellcolor{cyan!25}\textbf{46.67} & \cellcolor{cyan!25}\textbf{46.88} & \cellcolor{cyan!25}\textbf{36.00} & \cellcolor{cyan!25}\textbf{63.13} & \cellcolor{cyan!25}\textbf{63.13} & \cellcolor{cyan!25}\textbf{56.40} & \cellcolor{cyan!25}\textbf{62.59} & \cellcolor{cyan!25}\textbf{62.48} & \cellcolor{cyan!25}\textbf{55.80} \\
\midrule

\rowcolor{gray!20}\multicolumn{13}{c}{\textbf{Llama-3.2-1B}} \\
\textsc{No-RAG} & 32.69 & 24.15 & 14.94 & 1.33 & 2.68 & 0.00 & 25.20 & 17.84 & 12.40 & 19.74 & 14.89 & 9.11 \\
\textsc{Vanilla-RAG}           & 47.63 & 44.49 & 37.28 & 17.33 & 19.47 & 10.67 & 45.47 & 41.35 & 36.53 & 36.81 & 35.10 & 28.16 \\
\textsc{CoT}           & 26.33 & 3.85 & 0.15 & 0.00 & 1.63 & 0.00 & 16.27 & 6.75 & 1.00 & 14.20 & 4.08 & 0.38 \\
\textsc{CoN}           & 34.91 & 7.20 & 0.89 & 18.67 & 3.58 & 0.00 & 35.07 & 5.94 & 1.20 & 29.55 & 5.57 & 0.70 \\
\textsc{IRCoT}         & 49.85 & 44.40 & 35.80 & 21.33 & 23.79 & 14.67 & 45.00 & 44.27 & 37.40 & 38.73 & 37.49 & 29.29 \\
\textsc{DRAGIN}        & 36.24 & 36.64 & 34.62 & 6.67 & 8.37 & 4.00 & 21.60 & 20.19 & 17.00 & 21.50 & 21.73 & 18.54 \\
\textsc{SKR}           & 31.95 & 22.63 & 14.64 & 0.00 & 2.39 & 0.00 & 19.73 & 19.87 & 14.66 & 17.23 & 14.96 & 9.77 \\
\textsc{RAG-RL}        & 51.35 & 48.56 & 41.28 & 18.67 & 19.60 & 12.33 & 49.53 & 41.35 & 41.87 & 39.85 & 36.50 & 31.83 \\
\textsc{Search-R1}     & 45.52 & 42.62 & 35.39 & 15.33 & 17.52 & 8.67 & 42.73 & 41.35 & 33.40 & 34.53 & 33.83 & 25.82 \\
\textsc{ARMOR}~(warm-up stage)        & \cellcolor{cyan!10}67.01 & \cellcolor{cyan!10}66.80 & \cellcolor{cyan!10}64.35 & \cellcolor{cyan!10}30.67 & \cellcolor{cyan!10}35.19 & \cellcolor{cyan!10}24.00 & \cellcolor{cyan!10}42.53 & \cellcolor{cyan!10}41.39 & \cellcolor{cyan!10}34.20 & \cellcolor{cyan!10}46.74 & \cellcolor{cyan!10}47.79 & \cellcolor{cyan!10}40.85 \\
\textsc{ARMOR}     & \cellcolor{cyan!25}\textbf{74.11} & \cellcolor{cyan!25}\textbf{74.52} & \cellcolor{cyan!25}\textbf{72.78} & \cellcolor{cyan!25}30.67 & \cellcolor{cyan!25}33.08 & \cellcolor{cyan!25}\textbf{26.67} & \cellcolor{cyan!25}\textbf{50.07} & \cellcolor{cyan!25}\textbf{49.48} & \cellcolor{cyan!25}\textbf{42.00} & \cellcolor{cyan!25}\textbf{51.62} & \cellcolor{cyan!25}\textbf{52.36} & \cellcolor{cyan!25}\textbf{47.15} \\
\bottomrule
\end{tabular}
}
\caption{Performance comparing our method with baselines on three multi-hop QA datasets using Qwen-3-0.6B, Qwen-3-4B, Llama-3.2-1B as backbones. We report Correctness (Cor.), F1 score, and Exact Match (EM) in percentage (\%). The best results are highlighted in \textbf{bold}.}
\label{tab:our_results_appendix}
\end{table*}
\section{Additional Experimental Results}\label{app_sec:add_exp}
As shown in Table \ref{tab:our_results_appendix}, small-scale language models typically struggle with complex multi-hop reasoning and are highly susceptible to retrieval noise due to their limited instruction-following and reasoning capabilities.
For instance, on Qwen-3-0.6B, advanced RL-based methods like \textsc{Search-R1} collapse, achieving an average correctness of only 19.66\%, which is even lower than the \textsc{No-RAG} baseline (20.48\%).
This phenomenon indicates that standard RL optimization is unstable for weaker generic models. In stark contrast, \textsc{ARMOR} demonstrates exceptional efficacy on lightweight models. On Qwen-3-0.6B, \textsc{ARMOR} achieves an average correctness of 57.22\%, outperforming \textsc{IRCoT}.
Similarly, \textsc{ARMOR} surpasses \textsc{IRCoT} by 12.89\% in average correctness on Llama-3.2-1B. 
These results confirm that \textsc{ARMOR}'s adaptive curriculum learning acts as an effective \textit{teacher}, enabling even smaller \textit{student} generators to master robust reasoning capabilities that are otherwise inaccessible through standard training.

\begin{figure*}[h]
    \centering
    \includegraphics[width=\linewidth]{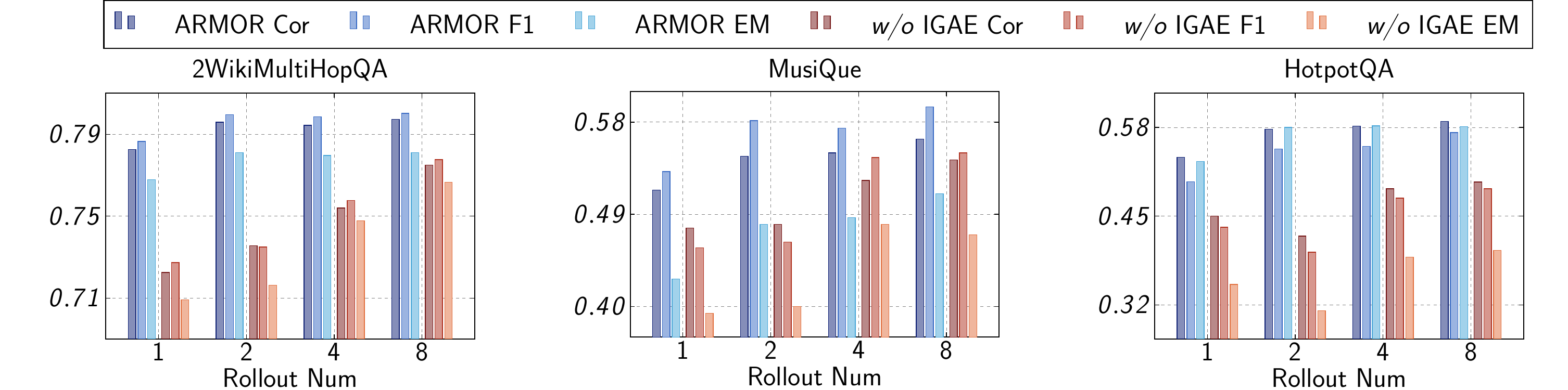}
    \caption{Sensitivity analysis of rollout number ($N$) on training stability using Qwen-3-1.7B.}
    \label{fig:Rollout-Num}
\end{figure*}

\subsection{Sensitivity Analysis}
\begin{figure}[H]
    \includegraphics[width=0.85\linewidth]{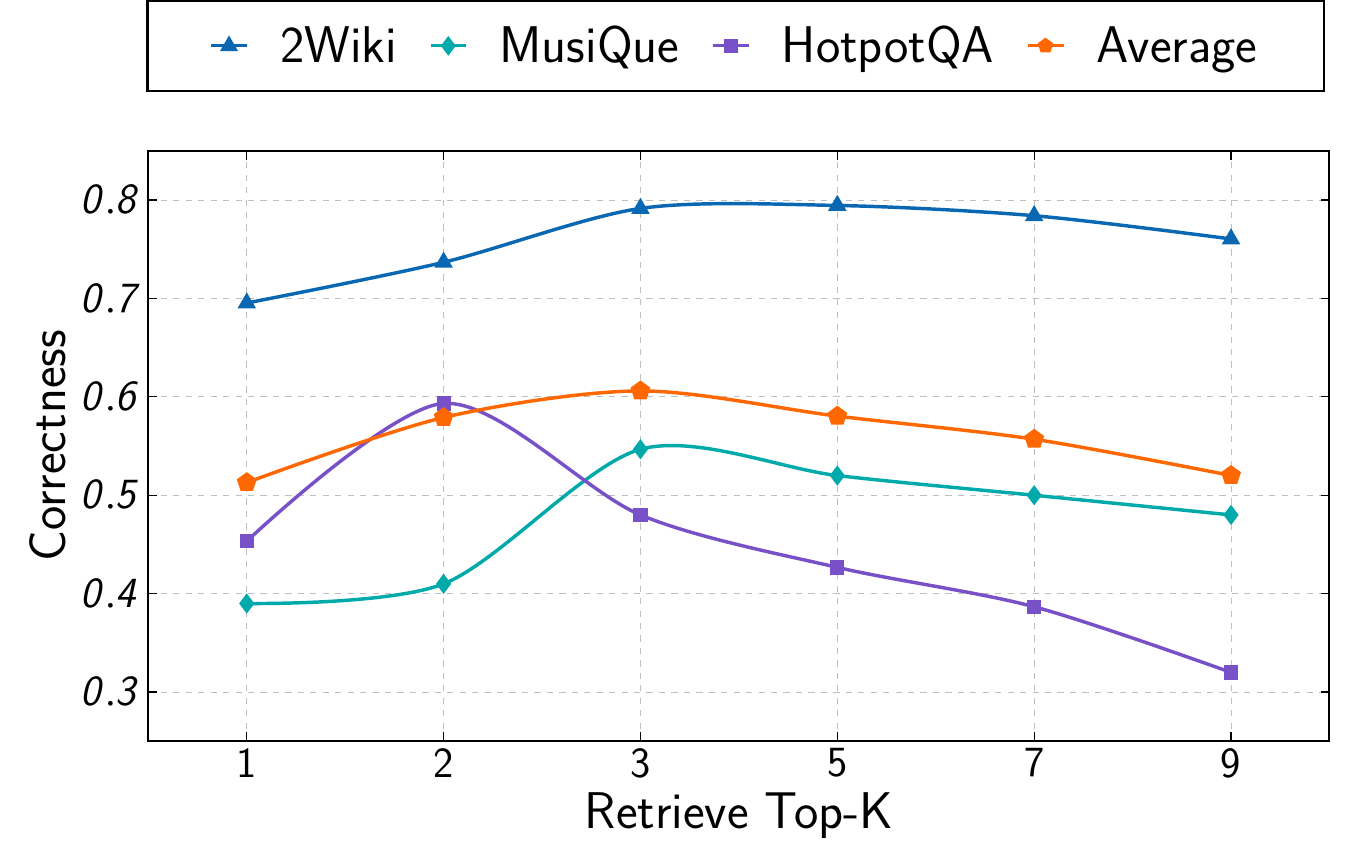}
    \caption{Impact of retrieval size ($K$) on generation correctness using Qwen-3-1.7B.}
    \label{fig:ret-topk}
\end{figure}

We conduct sensitivity analysis using Qwen-3-1.7B to evaluate ARMOR's stability under varying hyperparameter settings.
As illustrated in Figure~\ref{fig:ret-topk}, the generator's performance exhibits a trend of initially increasing and then decreasing as $K$~(the number of retrieved documents) increases.
Performance peaks at $K=3$, suggesting that a moderate amount of context provides sufficient information for reasoning.
However, further increasing does not yield continuous improvements.
This phenomenon indicates that retrieval is not ``the more, the better.'' 
Introducing an excessive number of documents significantly increases the risk of incorporating noise, which overwhelms the generator's reasoning capabilities and dilutes the signal from golden documents.

We analyze the impact of rollout size~($N$) on training stability and performance.
Figure~\ref{fig:Rollout-Num} compares ARMOR against the variant without Inter-Group Advantage Estimation (\textit{w/o IGAE}).
ARMOR exhibits remarkable sample efficiency, maintaining high correctness, F1, and EM scores even with a minimal rollout number ($N=1$).
In contrast, the \textit{w/o IGAE} variant is highly sensitive to $N$, performing poorly when $N$ is small.
With a small $N$ and the constrained output space of sub-tasks, the generated outputs within a single group often become identical.
By leveraging Inter-Group Advantage Estimation, ARMOR expands the comparison scope to the entire batch, ensuring effective advantage estimation regardless of group size.

\subsection{Case Study} \label{app_sec:case_study}

Table~\ref{tab:case-study} presents a concrete example of multi-hop question answering with different reasoning method. We illustrate how each method decomposes the question, interacts with retrieved evidence, and concludes a final answer. The case highlights distinct failure modes in \textsc{Search-R1} and \textsc{IRCoT}. In contrast, ARMOR demonstrates more structured reasoning and effective noise filtering from retrieved contexts.

\begin{table*}[htbp]
\centering
\caption{Comparative case study of \textsc{ARMOR}, \textsc{Search-R1}, and \textsc{IRCoT} on multi-hop QA with noisy retrieval.}
\label{tab:case-study}
\begin{tabularx}{\textwidth}{@{}lX@{}}
\toprule
\textbf{Question:} & When was Ole Rynning's father born? \\
\midrule
\textbf{Ground Truth:} & May 30, 1778 \\
\midrule
\textsc{\larger[0.7]\textbf{ARMOR:}} & 
\begin{minipage}[t]{\linewidth}
\think{To find out when Ole Rynning's father was born, I first need to identify who Ole Rynning's father is.} 
\subquestion{Who is Ole Rynning's father?} \search{True} \\
\doc{... He was the son of the priest Jens Rynning (1778–1857) ...} \\ \doctype{golden} \doc{...} \doctype{irrelevant} ...\\
\subanswer{Ole Rynning's father is Jens Rynning.} \\
\think{Now that I know Ole Rynning's father is Jens Rynning, I need to find his date of birth.} \\
\subquestion{When was Jens Rynning born?} 
\search{True} \\
\doc{Jens Rynning (May 30, 1778 – June 11, 1857) was a ...} \doctype{golden} \doc{Ole Rynning (April 4, 1809 – September 1838) was a ...} \\ \doctype{inconsequential} ...\\
\subanswer{Jens Rynning was born on May 30, 1778.} \\
\mainanswer{May 30, 1778.}
\end{minipage} \\

\midrule
\textsc{\larger[0.7]\textbf{Search-R1:}} & 
\begin{minipage}[t]{\linewidth}
\think{Okay, I need to find out when Ole Rynning's father was born. First, I should check if I have any existing information about Ole Rynning. I remember that Ole Rynning is a Danish footballer...} \\
\search{Ole Rynning birth date} \\
\info{... He was the son of the priest Jens Rynning (1778–1857) ...} \\
\search{Ole Rynning father birth date} \\
\info{... He was born in Stavanger as a son of Ole Bergesen ...} \\
\answer{Jens Rynning}
\end{minipage} \\

\midrule
\textsc{\larger[0.7]\textbf{IRCoT:}} & 
\begin{minipage}[t]{\linewidth}
\emph{After interleaving chain-of-thought reasoning with iterative retrieval, \\ the QA reader generates the final answer:}  \\
\doc{Jens Rynning (May 30, 1778 – June 11, 1857) was a ...} 
\\ \doc{Ole Rynning (April 4, 1809 – September 1838) was a ...}\\
\doc{... He was the son of the priest Jens Rynning (1778–1857) ...} \\ \doc{...} ...\\
\mainanswer{April 4, 1809.}
\end{minipage} \\
\bottomrule
\end{tabularx}
\end{table*}

\section{Details of LLM-Based Answer Evaluation}\label{app_sec:Details of Answer Evaluation}
\subsection{Prompts for Correctness Evaluation}
\begin{figure*}[h]
\centering
\setstretch{1.2}

\begin{tcolorbox}[width=0.9\textwidth, title=Prompt for the Evaluation of Correctness,
toptitle={0.3em}, bottomtitle={0.1em}]

You are an evaluator tasked with scoring the Candidate Answer.
\vspace{0.5em}

\textbf{Instructions:}

1. Compare the Candidate Answer to the Correct Answer in the context of the Question.

2. Assign a score from 0 to 5 based on accuracy and completeness:

\begin{enumerate}[leftmargin=1em, itemindent=0em, label={}, topsep=0.5em, itemsep=-0.3em]
   \item \emph{Score 0:} Completely incorrect or irrelevant.
   \item \emph{Score 1:} Partially correct but contains significant errors.
   \item \emph{Score 3:} Moderately correct but lacks some details or precision.
   \item \emph{Score 5:} Fully correct and matches the Correct Answer.
\end{enumerate}

3. Output ONLY the score as a single number (e.g. ``3''). Do not include any explanations.
\vspace{0.5em}

\textbf{Question:} \texttt{<}\emph{question}\texttt{>}

\textbf{Correct Answer:} \texttt{<}\emph{correct answer}\texttt{>}

\textbf{Candidate Answer:} \texttt{<}\emph{llm answer}\texttt{>}

\end{tcolorbox}
\caption{Prompt for the evaluation of correctness.}
\label{fig:Prompt for the evaluation of correctness}
\end{figure*}

In addition to traditional lexical metrics such as Exact Match and F1, we employ an LLM-based scoring approach using \texttt{GPT-5} to evaluate the answer correctness in RAG systems. 
As illustrated in Figure~\ref{fig:Prompt for the evaluation of correctness}, our evaluation prompt instructs the model to assess the factual accuracy and completeness of generated answers by comparing them against reference answers. 
This approach overcomes the limitations of surface-level matching inherent in traditional QA metrics. 
Moreover, we also leverage this LLM-based evaluation to compute rewards for robust reasoning task during RL training, providing more informative feedback to guide the model compared with traditional lexical metrics.

\subsection{Human AI Alignment}

\begin{figure}[H]
    \centering
    \includegraphics[width=0.9\linewidth]{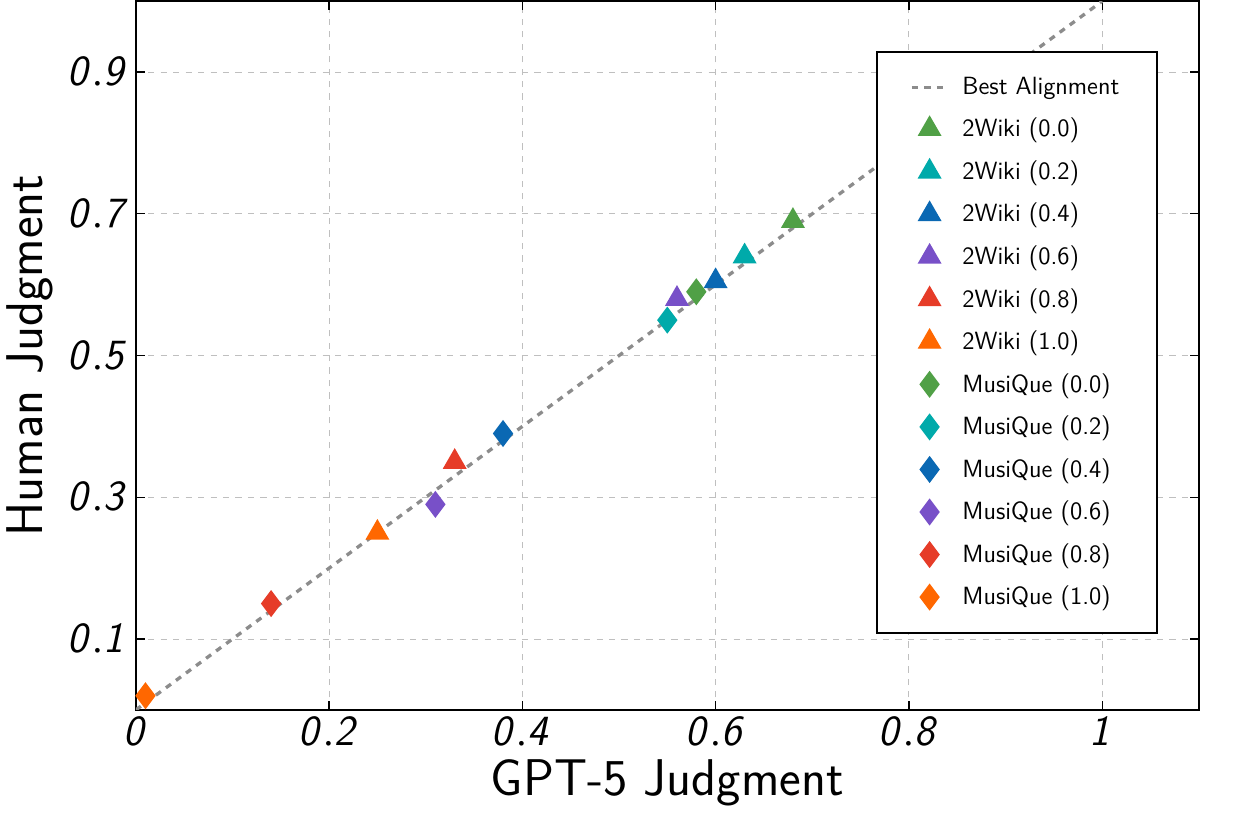}
    \caption{Comparison of correctness based on (1) Human and (2) \texttt{GPT-5} Judgement. Legend entries such as ``\textit{2Wiki} (0.6)'' denote results of using \textit{2WikiMultiHopQA} with 60\% noise ratio.}
    \label{fig:Human-AI Alignment}
\end{figure}

To validate the alignment between LLM-based automatic evaluation and human judgment, we conducted an alignment analysis on the performance of Qwen-3-1.7B at different noise ratios using \textsc{VanillaRAG} as the framework. 
We randomly sampled 100 instances from \textit{2WikiMultiHopQA} and \textit{MuSiQue} at different noise ratios. 
For each instance, we computed the average correctness assigned independently by human evaluators and \texttt{GPT-5}.
As shown in Figure~\ref{fig:Human-AI Alignment}, the scores exhibit high similarity, indicating strong alignment between LLM-based and manual judgements. 
These findings support the effectiveness of the LLM-based evaluation framework as a reliable surrogate for human judgment in large-scale experimental settings.

\end{document}